\documentclass[journal]{IEEEtran}

\usepackage{xcolor,soul,framed} 

\colorlet{shadecolor}{yellow}
\usepackage[pdftex]{graphicx}
\graphicspath{{../pdf/}{../jpeg/}}
\DeclareGraphicsExtensions{.pdf,.jpeg,.png}

\usepackage[cmex10]{amsmath}
\usepackage{array}
\usepackage{mdwmath}
\usepackage{mdwtab}
\usepackage{eqparbox}
\usepackage{url}
\usepackage[T1]{fontenc} 
\usepackage{amsmath}
\usepackage[cmintegrals]{newtxmath}
\usepackage{bm} 
\usepackage{comment}

\begin{document}
\noindent\fbox{%
    \parbox{\textwidth}{%
© 2021 IEEE.  Personal use of this material is permitted. Permission from IEEE must be obtained for all other uses, in any current or future media, including reprinting/republishing this material for advertising or promotional purposes, creating new collective works, for resale or redistribution to servers or lists, or reuse of any copyrighted component of this work in other works.}}

\bstctlcite{IEEEexample:BSTcontrol}
    \title{On the Adaptation of an Existing AUV into a Dedicated Platform for Close Range Imaging Survey Missions}
  \author{Yevgeni~Gutnik, Aviad~Avni, Tali~Treibitz,
      and Morel~Groper.\\



  \thanks{This work was supported by the The Leona M. and Harry B. Helmsley Charitable Trust, The Maurice Hatter Foundation, The Murray Foundation, the Israel Ministry of National Infrastructures, Energy and Water Resources Grant 218-17-008, and the Israel Ministry of Science, Technology and Space Grant 3-12487.} 
 }

\maketitle

\begin{abstract}
 Based on the need for high-resolution underwater visual surveys, this study presents the adaptation of an existing SPARUS II autonomous underwater vehicle (AUV) into an entirely hovering AUV fully capable of performing autonomous, close range imaging survey missions. This paper focuses on the enhancement of the AUV’s maneuvering capability (enabling improved maneuvering control), implementation of an state-of-the-art  thruster allocation algorithm (allowing optimal thrusters allocation and thrusters redundancy), and the development of an upgraded path-following controller to facilitate precise and delicate motions necessary for high resolution imaging missions. To facilitate the vehicle’s adaptation, a dynamic model is developed. The calibration process of the dynamic model coefficients initially obtained using well-accepted formulas, by computational fluid dynamics and in real sea experiments is presented. The in-house development of a pressure resistant imaging system is also presented. This system which includes a stereo camera and high-power lightning strobes was developed and fitted as a dedicated AUV payload.  Finally, the performance of the platform is demonstrated in an actual seabed visual survey mission.

\end{abstract}

\begin{IEEEkeywords}
\hl{dynamics of underwater vehicles, hovering Autonomous underwater vehicle (AUV), hydrodynamic coefficients, path following, thruster allocation, visual survey.  }
\end{IEEEkeywords}

%
\IEEEpeerreviewmaketitle


\section{Introduction}

\IEEEPARstart VISUAL SURVEYS of the seafloor enable us to quantify underwater benthic communities~\cite{pizarro2017simple},~\cite{tolimieri2008evaluating}, explore archaeological sites~\cite{gracias2013mapping}, inspect sub-sea structures~\cite{albiez2015flatfish}, map the seabed ~\cite{vidal2018online},~\cite{ludvigsen2007applications} and hence are extremely important in marine research. Today’s methods allow high-resolution and color 3D reconstructions of surveyed sites~\cite{jakuba2010high},~\cite{kim2009toward}. The underwater medium, however, presents significant challenges to imaging survey platforms. The rapid attenuation of light and the presence of backscatter limit the imaging range to only a few meters. Consequently, to map a large area, sets of images must be collected and stitched together into a photomosaic ~\cite{pizarro2003toward},~\cite{prados2012novel}.

\begin{figure}[t]
 \frame{\includegraphics[width=3.5in]{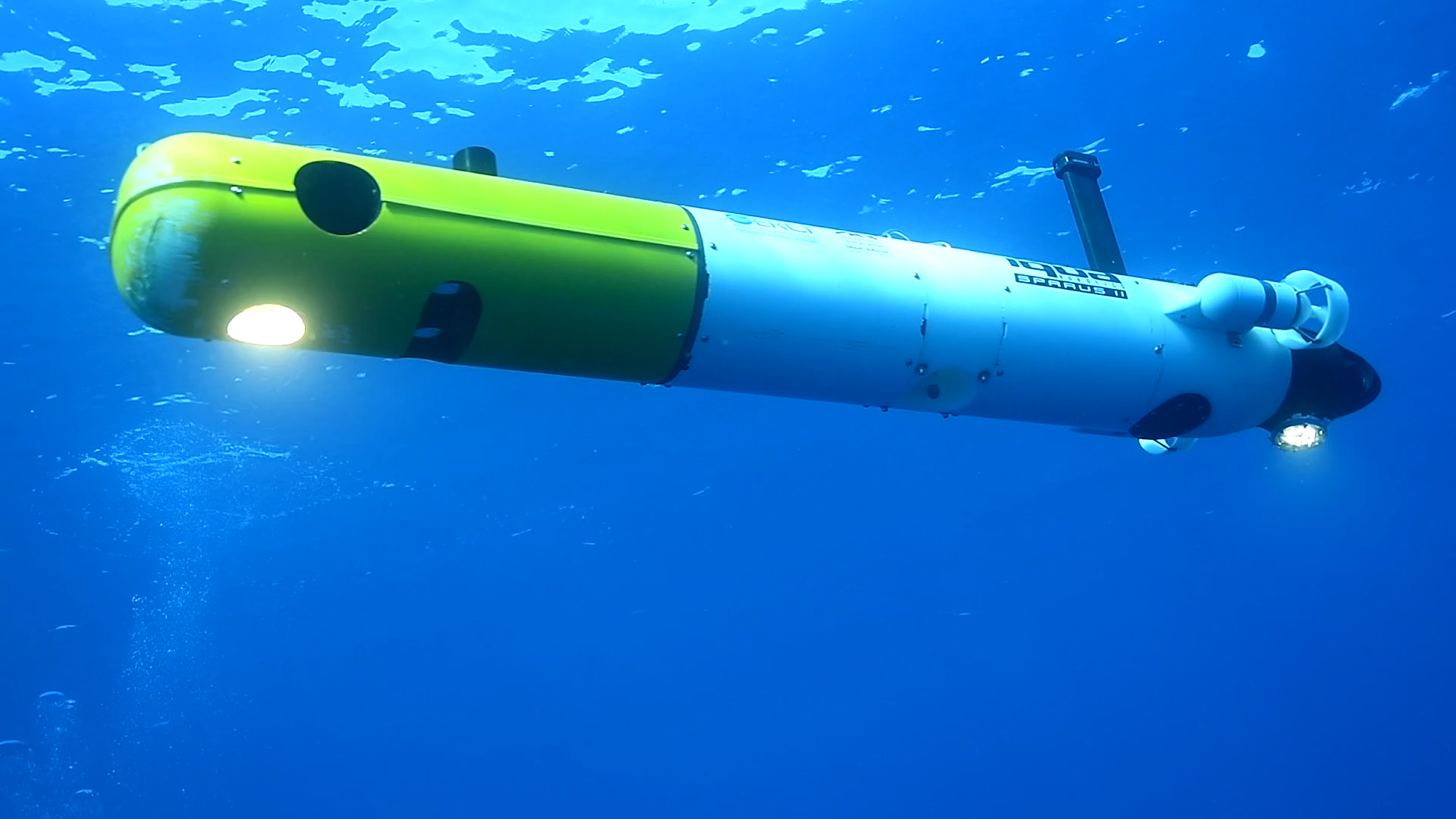}}
  \caption{ALICE AUV during sea experiments in the Mediterranean sea.}
  \label{alice_sea}
\end{figure}

The quality of a photomosaic depends on the ability to acquire images at constant altitude and orientation with respect to the seabed, on the amount of overlap between the images in the set ~\cite{ludvigsen2007applications},~\cite{pyo2015development} and on the accuracy of the underwater navigation. Acquiring such an image set requires roaming the camera with high precision along the mapping transects while maintaining constant altitude and orientation.

Often, image sets are collected by scuba divers; however, divers are limited by dive time and depth. In addition, they are unable to precisely navigate long transects while maintaining the camera at a constant orientation and altitude. The development of advanced underwater platforms such as remotely operated vehicles (ROVs) and autonomous underwater vehicles (AUVs) allows researchers to execute complicated missions while extending the effective time and depth, and eliminating the inherent risks of human involvement.

ROVs are generally powered and remotely controlled from surface support vessels via a tether that provides power, real-time control and video stream channels. The tether, however, limits the range and mobility of the vehicle. Furthermore, the deeper the missions are, the higher the deployment complexity is ~\cite{gracias2013mapping}.

AUVs are untethered, self-controlled vehicles, able to operate for substantial durations of time without returning to the surface~\cite{bewley2015australian}. Because they are untethered, AUVs can be deployed easily from relatively small vessels, which, therefore, increases flexibility and reduces operational costs. The untethered configuration enables AUVs to maneuver freely without the risk of tether entanglement, in particular at low altitudes and in complex environments. These features make AUVs an excellent candidate platform for underwater visual survey missions ~\cite{tolimieri2008evaluating,houts2012aggressive}. 
Typically, the required resolution for biodiversity analysis is in the order of  $1$ mm. For image resolutions of several megabytes the obtained field-of-view (FOV) is just a few square meters. In addition, water turbidity forces the imaging to be at close range. For high-quality photomosaics, high overlap between the images is necessary. Consequently, to enable collection of high-quality image sets, AUVs conducting photographic survey missions are required to operate with high precision at low speeds and at low altitudes. Achieving such precision is challenging, particularly in adverse sea conditions, in shallow waters and over rugged terrains.

Recently, special attention has been given to the development of AUVs capable of performing diverse missions, both at relatively high speeds for surveying large areas and at low speeds while hovering for close-range inspection of specific points of interest ~\cite{helgason2012low}. Usually, this new class of vehicles is characterized by having a single streamlined, torpedo-shaped hull with a rear propeller and control surfaces for high-speed missions, and multiple thrusters, typically mounted in a tunnel configuration, to provide precise maneuvering capabilities at low speed~\cite{packard2010hull},~\cite{philips2013delphin2},~\cite{wirtz2016iceshuttle}. While the configuration of tunnel thrusters offers reduced drag, it may require extension of the hull to physically accommodate the thrusters. Alternatively, the thrusters may be mounted externally to the hull with fixed mountings ~\cite{wang2019maneuverability} or pivoted mechanisms~\cite{allotta2016archaeology} to provide improved maneuverability and robustness in case of thruster failure.

The present work presents a process for upgrading an existing AUV with limited maneuverability and transforming it into a platform capable of performing delicate and precise maneuvers for close-range imaging missions. The maneuverability of the vehicle is improved by the addition of two lateral thrusters, providing sway and decoupled yaw control. To study and evaluate the enhanced propulsion system and calibrate the vehicle’s propulsion control system, a dynamic model is developed and validated with experimental data. A thruster allocation algorithm is developed to optimize the new, over-actuated propulsion system configuration and to provide redundancy in case of thruster failure. To facilitate precise and decoupled motion control along the imaging transects, an improved path-following controller is developed and integrated into the vehicle’s auto pilot system. An imaging payload, consisting of stereo, downward-looking cameras and a system of strobes is developed and integrated into the vehicle. Finally, the upgraded platform named “ALICE” (see Fig.~\ref{alice_sea}) is tested in a real underwater visual survey mission, where a substantial improvement in the vehicle’s motion control and the accuracy of its path following are demonstrated.

The paper is organized as follows: Section~\ref{sparus_auv} describes the basic configuration of the existing SPARUS II AUV while Section~\ref{alice_auv} describes the configuration of the modified platform “ALICE”, including the development of the dedicated visual imaging payload and the upgraded propulsion system. Section~\ref{auv_dynamics} describes the development of a six degrees-of-freedom (DOF) dynamic model for ALICE. ALICE’s hydrodynamic parameters and propulsion system model are described in Section~\ref{auv_coefficients}. Model validation with experimental, real sea trial data is described in Section~\ref{Model_validation}. The optimized thruster allocation algorithm is described in Section~\ref{thruster_alocation} while the improved path-following controller is described in Section \ref{path_controller}. Preliminary experiments using the new vehicle in imaging missions are described in Section \ref{imaging_experiments} and a demonstration of a complete imaging survey mission is presented in Section \ref{path_experiments}. Concluding remarks are presented in Section \ref{Conclusions}.

\section{The SPARUS II AUV}
\label{sparus_auv}

The SPARUS II~\cite{carreras2018sparus} is a torpedo-shaped AUV with partial hovering capabilities, developed by \emph{IQUA Robotics Inc\footnote{\url{http://iquarobotics.com/}}}. The vehicle’s operating speed is 0–2 m/s and its maximum operating depth is 200 m. Its small dimensions (\mbox{1.6 m} in length and 230 mm in diameter) and low weight (approx. 52 kg), allows its deployment from and retrieval to relatively small support surface vessels. For underwater navigation, the vehicle is equipped with an inertial measurement unit (IMU), pressure sensors, an acoustic Doppler velocity logger (DVL) and an ultra-short baseline localization modem (USBL). The control and communication architecture is based on open source software implemented in a ROS environment~\cite{quigley2009ros}. The payload volume can be configured to support payloads with a maximum capacity of 8 liter and weight in air of 7 kg.

The basic configuration of the SPARUS II AUV employs three thrusters to control the AUV in 3-DOF. Two horizontal thrusters mounted at the stern control surge motion by creating equal thrust, and yaw motion, by creating differential thrust. A single vertical tunnel thruster, located at the vehicle’s center of buoyancy, controls heave motion.
The sway, roll and pitch motions are not actively controlled and, therefore, the vehicle is considered as under-actuated. Consequently, compensation for lateral drift, which may occur due to environmental disturbances, is indirectly controlled by employing the horizontal thrusters. This method, however, results in side-slip~\cite{xia2019improved}, causing slow and inaccurate response to unpredictable perturbations.

\section{Modified SPARUS II AUV - ``ALICE"}
\label{alice_auv}

\begin{figure}[t]
  \begin{center}
  \includegraphics[width=3.5in]{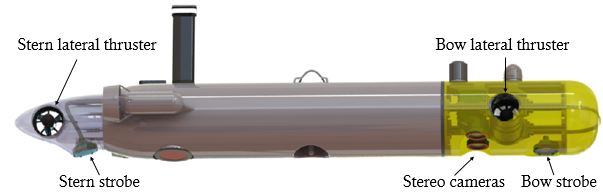}
  \caption{ALICE AUV – An autonomous platform for acquiring high resolution 3D photomosaics. The wet payload section consists of the nose lateral thruster, the stereo cameras and the nose strobe. The stern was redesigned to fit a second lateral thruster and the tail strobe. A GPU, LED drivers and a micro-controller to trigger the strobes were fitted in the vehicle’s dry section.} 
\label{alice_auv_cad}
  \end{center}
\end{figure}

\subsection{Enhancement of the  Propulsion System}
To provide uncoupled sway control and improve yaw motion control, the existing propulsion system was upgraded by the addition of two \emph{Blue Robotics  \footnote{\url{https://bluerobotics.com/}}} T-200 lateral thrusters. To gain maximum yaw moment, the thrusters are located at the farthest possible distance from the center of buoyancy. The forward thruster, located in the AUV’s nose, is installed inside a tunnel, in the wet payload section, while the tail thruster is installed inside a new wet tail cone section, as illustrated in Fig.~\ref{alice_auv_cad}. This section was completely redesigned to accommodate the tail lateral thruster and a strobe light. The thrusters’ power cables are connected to dedicated drivers located in the AUV’s dry pressure-resistant section through waterproof connectors (manufactured by \emph{SubConn Inc\footnote{\url{https://www.macartney.com/what-we-offer/systems-and-products/connectors/subconn/}}}). The location of the thrusters in the vehicle’s wet sections allows easy maintenance and eliminates the need to redesign the vehicle’s dry section. The tunnel thruster configuration preserves the streamlined shape of the hull, resulting in a relatively small drag penalty. In addition, this design has minimal impact on the vehicle’s launch and recovery operations.

\subsection{Development of a Dedicated Imaging Payload}
\label{imaging_payload}

\begin{figure}[t]
  \begin{center}
  \includegraphics[width=1.0\linewidth]{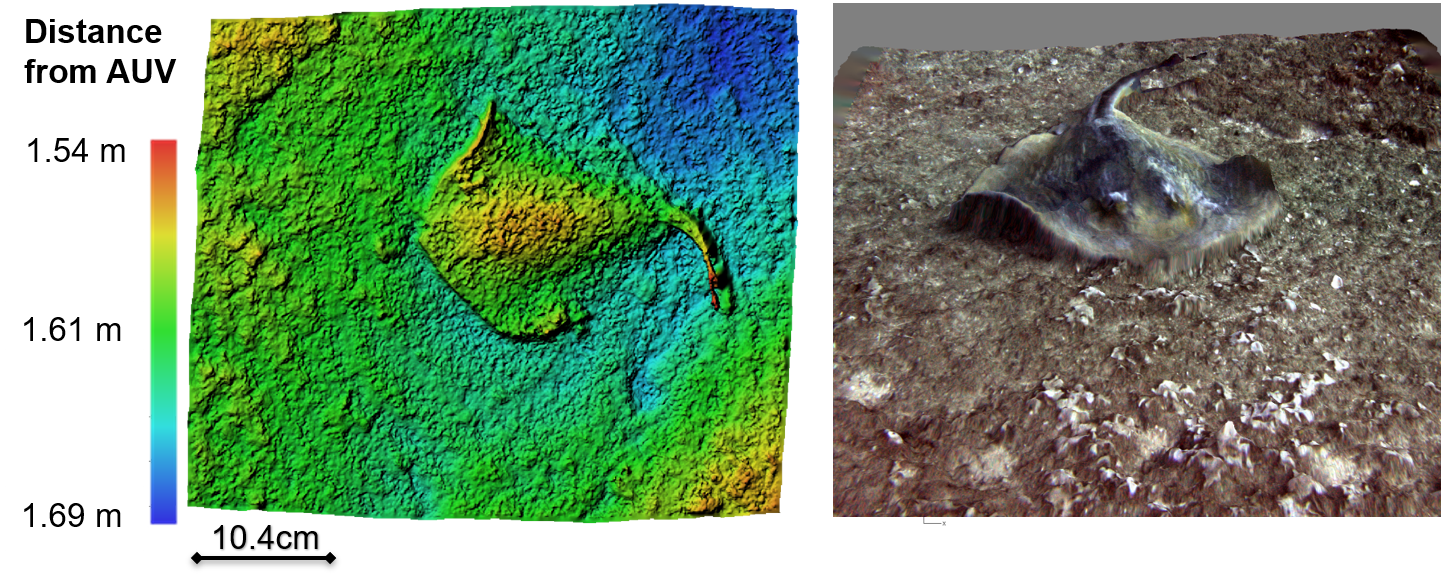}
  \caption{3D reconstruction using the stereo pair of cameras. [Left] Altitude of AUV from the stingray. [Right] Front view of the 3D model generated from the Alice AUV stereo system. Image resolution and field of view at 2 m altitude for two optional lenses. Values for an underwater flat port housing are calculated.}\label{ray 3d}
  \end{center}
\end{figure}

The in-house designed stereo imaging system developed to address the need to create high-resolution  ($\sim1$ mm) 3D photomosaics of the seafloor is illustrated in Fig.~\ref{alice_auv_cad}. The system was designed to acquire images from an altitude of 2 m, considered as the safety altitude for the AUV. Two 9.2 Megapixel \emph{Allied Vision\footnote{\url{https://www.alliedvision.com/}}} Manta-G917 cameras were selected to provide high-resolution images and installed in a compact pressure-resistant, in-house developed, package designed to fit the vehicle’s payload bay. To avoid distortion that may be caused by the camera housing’s flat port~\cite{treibitz2011flat}, the system was designed to support lenses that provide a field of view of up to $70^\circ$. The characteristics of the two lenses that were considered are summarized in Table~\ref{table:FOV}. For the purpose of this study, $12.5$ mm lens were selected. The stereo pair is installed at a baseline of $100$ mm apart, which is the maximum distance the payload section supports. A shorter baseline would have resulted in reduced accuracy of the 3D reconstruction. An example is shown in Fig.~\ref{ray 3d}, which demonstrates a 3D reconstruction of a stingray using the stereo pair of cameras in a single location.
For acquiring images at low light scenes, two high-power LED strobes in in-house designed housings were developed. Each strobe consists of an array of 20 Cree XHP-35 LEDs, providing $30,000$ lumens at $4000$K color temperature. The beam angle is $120^\circ$, fully covering the camera’s FOV.
 
 \begin{table}[t]
\begin{center} 
\begin{tabular}{|l|c|c|}
	\hline
	Focal length  & 12.5 mm & 25 mm \\
	\hline
    Horizontal FOV  & 1.4 m &  0.75 m \\
	\hline
    Vertical FOV  & 1.16 m &  0.6 m\\
	\hline    
     Resolution   & 0.45 mm &  0.22 mm\\
	\hline   
\end{tabular}
   \caption{Image Resolution and Field of View at $2$ m Altitude for Two Optional Lenses. Values are Calculated for a Flat Port Housing Underwater.}
 \label{table:FOV}
 \end{center}
\end{table}	
The stereo imaging system and one strobe were installed in the nose wet payload section while the second strobe was installed in the tail cone section (see   Fig.~\ref{alice_auv_cad}). A custom-made Arduino based microcontroller was developed to control the strobes’ intensity and synchronize the cameras and the strobes, triggering signals at a frame rate of up to 10 fps. An \emph{NVIDIA\footnote{\url{https://www.nvidia.com/}}} Jetson TX2 GPU was used for image acquisition, camera control and communication with the navigation system. Both the microcontroller and the Jetson were installed in the main hull and connected to the cameras and strobes through underwater cables and SubCon pluggable connectors.

\section{Dynamic Modeling of ALICE AUV}
\label{auv_dynamics}
To support the development of the upgraded propulsion system and optimize the performance of the vehicle’s propulsion control in various scenarios of underwater imaging survey missions, a detailed dynamic model was developed and implemented in a numerical scheme. The model computes the dynamic response of the vehicle, in 6-DOF, to control forces and to environmental disturbances. Special attention was given to the vehicle’s response during delicate maneuvers at slow speeds, necessary for close to seabed imaging missions.


\subsection{Coordinate Systems and Kinematic Equations of Motion}

\begin{figure}[t]
  \begin{center}
  \includegraphics[width=3.5in]{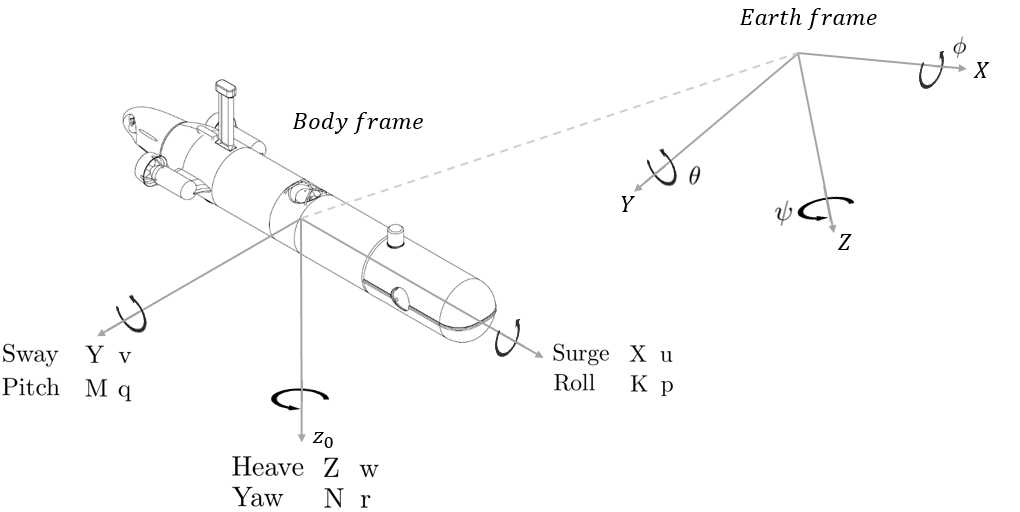}
  \caption{Body-fixed and Earth-fixed reference frames.}\label{frames}
  \end{center}
\end{figure}

The motion of the vehicle in 6-DOF is described by employing the \emph{SNAME} notation~\cite{fossen2011handbook} in the body-fixed and the Earth-fixed reference frames, as detailed in Table~\ref{sname} and illustrated in Fig.~\ref{frames}. The body-fixed frame is employed to describe the accelerations, velocities, forces and moments acting on the vehicle with the origin fixed at the vehicle’s center of buoyancy, the x-axis extending along the hull axis of symmetry directing toward the nose, the y-axis directing toward the starboard side, and the z-axis directing downwards. The Earth-fixed frame employs the North-East-Down (NED) convention to describe the vehicle’s trajectory and orientation.

\begin{table}[t]
	\begin{tabular}{ |l|l|p{0.8cm}|p{0.8cm}|p{0.8cm} c | }
		\hline	
		DOF & Description  & Forces and moments & Linear and angular velocity & Position and Euler angles& \\ \hline
		Surge & Motion   in    the x-direction & \centering{X} & \centering{u} & \centering{x}&    \\
		Sway  & Motion   in    the y-direction & \centering{Y} & \centering{v} & \centering{y}&    \\ 
		Heave & Motion   in    the z-direction & \centering{Z} & \centering{w} & \centering{z} &    \\
		Roll  & Rotation about the x-axis      & \centering{K} & \centering{p} & \centering{$\phi$} &  \\
		Pitch & Rotation about the y-axis      & \centering{M} & \centering{q} & \centering{$\theta$}& \\
		Yaw   & Rotation about the z-axis      & \centering{N} & \centering{r} & \centering{$\psi$} &  \\ 
		\hline
			\end{tabular}
	\caption{SNAME Notation for Marine Vehicles}\label{sname}
\end{table}

The relation between the body-fixed and the Earth-fixed frames is given by:
\begin{equation}\label{transformation}
\dot{\Vec{\eta}} = J(\Vec{\eta})\Vec{\nu} 
\end{equation}
where $\eta =[x,y,z,\psi,\theta,\phi]^T$ represents the generalized position and orientation of the vehicle with respect to the Earth-fixed frame, $\nu=[u,v,w,p,q,r]^T$ represents the linear and angular velocities in the body-fixed frame and $J(\eta)=[J_1,J_2]^T$ is the transformation matrix between the body-fixed and the Earth-fixed frames as presented in ~\cite{fossen2011handbook}. 

\subsection{Dynamic Equations of Motion}
The dynamic motion of the vehicle is modeled using the generalized form of the equations of rigid body dynamics as specified by Fossen~\cite{fossen2011handbook}. Employing the \emph{SNAME} notation and considering the forces and moments action on a fully submerged vehicle, the nonlinear equations of motion in the body-fixed frame are expressed by:
\begin{equation}\label{general_form}
M\dot{\Vec{\nu}} + C(\Vec{\nu})\Vec{\nu} + D(\Vec{\nu})\Vec{\nu} + \Vec{g}(\Vec{\eta}) = \Vec{\tau_c}
\end{equation}
where $M$ is the system inertia matrix (including the rigid body $M_{RB}$ and the added mass  $M_{A}$ matrices); $C(\Vec{\nu})$ is the Coriolis/centripetal matrix consisting of the rigid body matrix $C_{RB}(\Vec{\nu})$ and the added mass matrix $C_{A}(\Vec{\nu})$; $D(\Vec{\nu})$ is the hydrodynamic damping matrix combining the nonlinear damping matrix and the body lift matrix; $\Vec{g}(\Vec{\eta})$ is the  restoring forces and moments vector and $\Vec{\tau_c} =[X,Y,Z,K,M,N]^T$ is the control forces and moments vector created by the vehicle's thrusters. 

\section{Identification of the Hydrodynamic Coefficients}
\label{auv_coefficients}
The hydrodynamic forces and moments acting on the vehicle are described in terms of hydrodynamic coefficients (derivatives). In this work, the vehicle’s coefficients are assumed to be constant and the following assumptions are made:
\begin{enumerate}
    \item The vehicle is completely submerged and operates sufficiently deep to allow us to disregard the influence of sea surface and waves.
    
    \item The hydrodynamic effects due to interaction between the vehicle components are small such that the coefficients of each component may be estimated individually. Consequently, to estimate coefficients of the entire vehicle analytically, the principle of superposition may be employed.
    
    \item The vehicle posses port-starboard symmetry.
    
    \item The vehicle’s components considered in the computation of the hydrodynamic coefficients are the hull, thrusters and mast with other small appendages disregarded. 
    
\end{enumerate}
The hydrodynamic coefficients are estimated by semi-analytical, empirical, and computational (CFD) methods. Later, as part of the sea trials, these coefficients are validated and calibrated accordingly. The coefficients as estimated and employed in the dynamic model are summarized in Table  ~\ref{coeff_results}. 


\subsection{Added Mass}
The axial added mass of the hull, $X_{\dot{u}}^{\rm hull}$ and of the horizontal thrusters, $X_{\dot{u}}^{\rm th}$ are estimated by approximating their shapes to ellipsoids and employing the following empirical formula~\cite{fossen2011handbook}:
\begin{equation}\label{axial_added_mass}
X_{\dot{u}}^{\rm hull} = - \frac{\alpha_0}{2 - \alpha_0}m_e
\end{equation}
where $m_e$ is the mass of the ellipsoid, given by:
\begin{equation}\label{axial_added_me}
m_e = \frac{4}{3}\pi\rho\frac{l}{2}\Big(\frac{d}{2}\Big)^2    
\end{equation}
The parameter $\alpha_0$ is determined by:
\begin{equation}\label{axial_added_a0}
\alpha_0= \frac{2(1-e^2)}{e^3}\Big(\frac{l}{2}ln\frac{1+e}{1-e}-e\Big)    
\end{equation}
with the eccentricity $e$  defined as:
\begin{equation}\label{axial_added_e}
e = 1 - \Big(\frac{d}{l}\Big)^2   
\end{equation}

\begin{figure}[t]
  \includegraphics[width=3.5in]{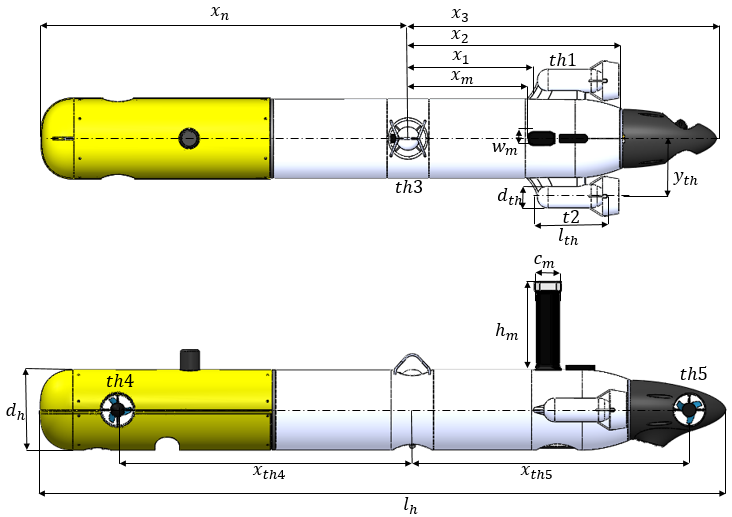}
  \caption{ALICE AUV dimensions}\label{sparus_dimensions}
\end{figure}

$X_{\dot{u}}^{\rm hull}$ is obtained by substituting the major axis $l$ with $l_h$ and the minor axis $d$ with $d_h$. $X_{\dot{u}}^{\rm th}$ is obtained by substituting the major axis $l$ with $l_{\rm th}$ and the minor axis $d$ with $d_{\rm th}$, where $l_h, l_{\rm th}, d_h, d_{\rm th}$ are illustrated in Fig.~\ref{sparus_dimensions}.

The axial added mass coefficient of the mast, denoted by $X_{\dot{u}}^{\rm mast}$, is determined by employing the empirical expression ~\cite{hagist1965experimental}, presented as a rod with an elliptic cross-section:
\begin{equation}\label{mast_axial_added_mass}
X_{\dot{u}}^{\rm mast} = - \frac{1}{4}\pi\rho (w_{m})^2 h_{m} 
\end{equation}

where $w_{m}$ and $h_{m}$ are the width and height of the mast, respectively, as illustrated in Fig.~\ref{sparus_dimensions}. The vehicle’s total axial added mass coefficient is obtained by summation of the components of the axial added mass coefficients, i.e., hull, thrusters~(\ref{axial_added_mass}) - (\ref{axial_added_e}) and  mast~(\ref{mast_axial_added_mass}):

\begin{equation}\label{x_u_dot}
X_{\dot{u}} = X_{\dot{u}}^{\rm hull} + 2 \cdot X_{\dot{u}}^{\rm th} + X_{\dot{u}}^{\rm mast}    
\end{equation}

Looking at the vehicle’s symmetry around the horizontal plane, the sole contribution to the added mass pitching moment is related to the mast. Consequently, the coefficient $M_{\dot{u}}$ is estimated by:

\begin{equation}\label{mast_axial_added_mass_moment}
M_{\dot{u}}^{\rm mast} =  X_{\dot{u}}^{\rm mast} \cdot \frac{d_{h} + h_{m}}{2} 
\end{equation}

The cross-flow added mass coefficients for the hull and the thrusters are estimated by employing the principle of strip theory. The bodies are divided into a finite number of 2D circular strips, where the added mass of a circular strip is given by~\cite{fossen2011handbook}:

\begin{equation}\label{cross_added_mass_ma}
m_a^c = - \pi \rho r(x)^2  
\end{equation}

The section of the hull that includes the mounts for the horizontal thrusters is modeled as a circular strip with fins with the added mass given by~\cite{fossen2011handbook}:
\begin{equation}\label{cross_added_mass_ma_f}
m_a^h  = - \pi \rho \Bigg[\frac{b^2-r(x)^2}{b^2}\Bigg]  
\end{equation}
where $r(x)$ is the local radius of the circular strip and b is the span of the mounts measured from the section’s center.

Consequently, the cross-flow added mass coefficients of the entire hull are given by:
\begin{align}
Y_{\dot{v}}^{\rm hull} =  &\int_{x_n}^{x_3}  m_a^c(x) dx \\
N_{\dot{v}}^{\rm hull} =  &\int_{x_n}^{x_3}  m_a^c(x)x dx \\
N_{\dot{r}}^{\rm hull} =  &\int_{x_n}^{x_3}  m_a^c(x)x^2 dx \\
Z_{\dot{w}}^{\rm hull} =  &\int_{x_n}^{x_1}  m_a^c(x) dx + \int_{x_1}^{x_2}  m_a^h(x) dx + \int_{x_2}^{x_{3}}  m_a^c(x) dx \\
M_{\dot{w}}^{\rm hull} =  &\int_{x_n}^{x_1}  m_a^c(x)x dx + \int_{x_1}^{x_2}  m_a^h(x) xdx + \\
&\int_{x_2}^{x_{3}}  m_a^c(x) xdx \\
M_{\dot{q}}^{\rm hull} =  &\int_{x_n}^{x_1}  m_a^c(x)x^2 dx + \int_{x_1}^{x_2}  m_a^h(x) x^2dx + \\
&\int_{x_2}^{x_{3}}  m_a^c(x) x^2dx 
\end{align}
and the cross-flow added mass coefficients of the horizontal thrusters are given by:
\begin{equation}\label{cross_flowthrusters}
Y_{\dot{v}}^{\rm th} = Z_{\dot{w}}^{\rm th}=  \int_{x_1}^{x_2}  m_a^c(x) dx 
\end{equation}
$x_1$, $x_2$, $x_3$ and $x_n$ are illustrated in Fig.~\ref{sparus_dimensions}.

The cross-flow added mass coefficient of the mast is estimated following the empirical expression presented in~\cite{hagist1965experimental} for a rod with an elliptic cross-section in cross flow. 

\begin{equation}\label{cross_flow_mast}
Y_{\dot{v}}^{m} =  - \frac{1}{4} \pi \rho c_m^2 h_m
\end{equation}

\subsection{Hydrodynamic Damping}

In the framework of the developed model and considering the assumptions listed above, the main components of the hydrodynamic damping are the quadratic terms of skin friction and vortex shading while coupled terms and terms higher than second order are omitted. The axial term of the hydrodynamic damping, ($X_{u|u|}$) is expressed by~\cite{horner1965fluid}:

\begin{equation}\label{axial_drag}
X_{u|u|} =  \frac{1}{2} \rho A C_{d_x} u |u|
\end{equation}

where $A$ is the frontal area and $C_{d_x}$ is the AUV’s combined drag coefficient computed by superposition of the specific drag forces for the AUV’s hull, horizontal thruster, and mast. For each component, the drag coefficient is estimated based on the component shape and the flow regime.

The axial drag coefficients for the hull and the horizontal thrusters are estimated by employing the empirically obtained equation~(\ref{axial_hull}), derived for ellipsoid bodies in axial flow~\cite{horner1965fluid}:
\begin{equation}\label{axial_hull}
C_{d_x} =  0.44 \bigg(\frac{d}{l}\bigg) + 4 C_f \bigg(\frac{l}{d}\bigg) + 4 C_f \bigg(\frac{d}{l}\bigg)^{\frac{1}{2}}
\end{equation}
where $C_f$ is the skin friction coefficient, following \emph{ITTC-57} correlation line~\cite{ittc2002guidelines}:
\begin{equation}\label{ittc_57_cf}
C_f =  - \frac{0.075}{(log_10R_e -2)^2}
\end{equation}

The cross-flow drag coefficients of the hull and the horizontal thrusters are estimated as in~\cite{uematsu1995effects} by:

\begin{equation}\label{cross_drag_hull}
C_{d_{cross}} =  C_{d0}k_1
\end{equation}
where $C_{d0} = 1.2$ is the drag coefficient of a 2D cylinder in cross flow and $k_1$ is determined by:
\begin{equation}\label{ideal_rectifier_resistance}
k_1 =
\begin{cases}
0.58 + 0.17 \bigg[ log_{10} \big(\frac{l}{d}\big) \bigg]^{1.6} & \frac{l}{d}  \leq 57.5 \\
    1                                                & \frac{l}{d} > 57.5
\end{cases}
\end{equation}

The drag coefficients of the mast and the horizontal thruster mounts, for both axial and cross-flow, are computed by employing drag coefficients of basic shapes provided in~\cite{horner1965fluid}, where the mast and the mounts are treated as rectangular plates. The drag coefficients of the thrusters’ tunnels are estimated following the method provided in~\cite{palmer2009analysis}.
\subsection{Body Lift}

When the AUV is flying at an angle of attack $\alpha$, or at a side-slip angle $\beta$, the flow may separate from the aft section, causing a pressure difference that creates lift force and moment. The  terms of the lift forces and moments are expressed by~\cite{hoerner1975fluid}:
\begin{equation}\label{lift__force_y}
Y = \frac{1}{2}\rho U^2 d_h^2 C_{L\beta}  \beta
\end{equation}
\begin{equation}\label{lift__force_general}
Z = \frac{1}{2}\rho U^2 d_h^2 C_{L\alpha}  \alpha
\end{equation}
\begin{equation}\label{lift__moment_general}
M = \frac{1}{2}\rho U^2 d_h^2 x_{\rm cp} C_{L\alpha}  \alpha 
\end{equation}
\begin{equation}\label{lift__moment_n}
N = \frac{1}{2}\rho U^2 d_h^2 x_{\rm cp} C_{L\beta}  \beta 
\end{equation}
where $U = \sqrt{ u^2 + w^2 }$, $C_{L\alpha}$ and $C_{L\beta}$ are the lift coefficients and $x_{\rm cp}$ is the center of pressure. Employing the approximation for small angles, $U \approx u$, angles $\alpha$ and $\beta$ may be rewritten as: 
\begin{equation}\label{approx_alpha}
\alpha = \tan^{-1}\frac{w}{u} \approx \frac{w}{u}
\end{equation}
\begin{equation}\label{approx_beta}
\beta = \tan^{-1}\frac{v}{u} \approx \frac{v}{u}
\end{equation}

Substituting (\ref{approx_alpha})-(\ref{approx_beta}) into (\ref{lift__force_y})-(\ref{lift__moment_n}) allows us to express the coefficients of lift forces and moments as follows:
\begin{equation}\label{lift_z}
Z_{uw}=-\frac{1}{2} \rho d_h^2 C_{L\alpha} 
\end{equation}
\begin{equation}\label{lift_y}
Y_{uv}=-\frac{1}{2} \rho d_h^2 C_{L\beta} 
\end{equation}
\begin{equation}\label{lift_m}
M_{uw} = Z_{uw} x_{\rm cp} 
\end{equation}
\begin{equation}\label{lift_n}
N_{uv} = Y_{uv} x_{\rm cp}
\end{equation}
where the lift coefficient $C_{L\alpha}$ and the center of pressure, $x_{\rm cp}$ are estimated by employing the empirical expressions (\ref{cl_alpha})-(\ref{x_cp}), presented in~\cite{hoerner1975fluid}:
\begin{equation}\label{cl_alpha}
C_{L\alpha} = 0.003 \bigg( \frac{180}{\pi} \bigg) \bigg( \frac{l_h}{d_h} \bigg)
\end{equation}
\begin{equation}\label{x_cp}
x_{\rm cp} = x_n - 0.65 \cdot l_h
\end{equation}

\begin{table}\centering
	\begin{tabular}{ |c|c|c|c| }
		\hline
		\centering
		Coefficient   & Value                                & Coefficient          & Value    \\  \hline
	    $X_{\dot{u}}$ & -2.806  $~[kg]$                      & $K_{\dot{p}}$        & -0.042 $~\big[\frac{kg \cdot m^2}{rad}\big]$     \\\hline
	    $Y_{\dot{v}}$ & -78.459 $~[kg]$                      & $X_{u|u}$            & -7.616    $~\big[\frac{kg}{m} \big]$             \\\hline
	    $Y_{\dot{r}}$ & -8.529  $~\big[\frac{kg \cdot m}{rad}\big]$ & $Y_{v|v|}$    & -214.398  $~\big[\frac{kg}{m}\big]$              \\\hline
	    $K_{\dot{v}}$ & 0.216   $~[kg \cdot m ]$             & $Z_{w|w|}$           & -214.398 $~\big[\frac{kg}{m}\big]$               \\\hline
	    $K_{\dot{p}}$ & -0.042  $~\big[\frac{kg \cdot m^2}{rad}\big]$ & $M_{w|w|}$  & 26.634   $~[kg]$                                 \\\hline
	    $N_{\dot{p}}$ & -0.102  $~\big[\frac{kg \cdot m^2}{rad}\big]$& $N_{v|v|}$   & -26.17  $~[kg]$                                 \\\hline
	    $Z_{\dot{w}}$ & -69.536 $~[kg] $                     & $K_{p|p|}$           & 0.192    $~\big[\frac{kg \cdot m^2}{rad^2}\big]$ \\\hline
	    $M_{\dot{u}}$ &  0.0321 $~[kg \cdot m ] $            & $K_{v|v|}$           & 3.397    $~[kg]$                                 \\\hline	  
	    $N_{\dot{v}}$ & -8.529  $~[kg \cdot m ] $            & $M_{q|q|}$           & -180.682 $~\big[\frac{kg \cdot m^2}{rad^2}\big]$ \\\hline	 
	    $M_{\dot{w}}$ & -11.253 $~[kg \cdot m ] $            & $N_{r|r|}$           & -180.381 $~\big[\frac{kg \cdot m^2}{rad^2}\big]$ \\\hline	
	    $M_{\dot{q}}$ & -20.963 $~\big[\frac{kg \cdot m^2}{rad}\big]$ & $M_{u|u|}$  & 0.144    $~[kg]$                                 \\\hline	 
	    $N_{\dot{r}}$ & -22.537 $~\big[\frac{kg \cdot m^2}{rad}\big]$ & $K_{r|r|}$  & 0.753    $~\big[\frac{kg \cdot m^2}{rad^2}\big]$ \\\hline		  
\end{tabular}
	\caption{Model Estimated Coefficients}
	\label{coeff_results}
\end{table}

\subsection{CFD Computation of Hydrodynamic Coefficients}

\begin{figure*}[t]
  \includegraphics[width=1\linewidth]{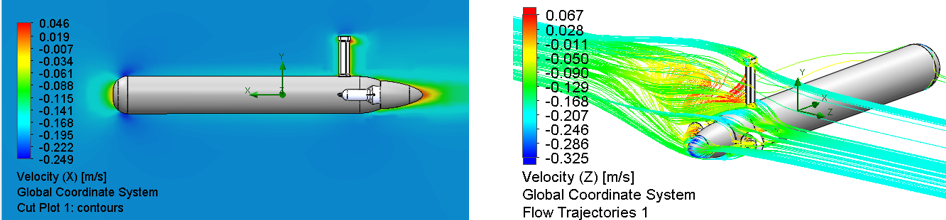}
  \caption{CDF simulation of the hull in axial (left) and cross flow (right) of 0.2 m/s.}
  \label{cfd}
\end{figure*}

In this work, the commercial \emph{Solidworks Flow Simulation package\footnote{\url{https://www.solidworks.com/product/solidworks-flow-simulation}}} was used to compute selected coefficients of drag and body lift with the goal of validating the results obtained through empirical expressions.
The computation of the drag coefficients for the complete vehicle, including the hull, the horizontal thrusters and their mounts and the mast, was performed at flow speeds of 0.2 m/s in axial and cross-flow directions as illustrated in  Fig.~\ref{cfd}. The generalized forces and moments  $[X,Y,Z,K,M,N]$ are defined as goal parameters and the coefficients were computed by dividing the computed forces and moments by the quadratic flow speed as, for instance, the axial drag coefficient is computed by:

\begin{equation}\label{cfd_coeff}
x_{u|u|} = \frac{X}{u|u|}
\end{equation}

For computation of the hull’s lift force and moment coefficients, two components of flow velocity were defined. The axial component was set to 0.25 m/s and the cross-flow component to 0.05 m/s, resulting in an angle of attack of approx.  $11^{\circ}$. Consequently, the lift force and moment coefficients were computed by:

\begin{equation}\label{cfd_lift_f}
Z_{uw} = \frac{Z}{uw}
\end{equation}
\begin{equation}\label{cfd_lift_m}
M_{uw} = \frac{M}{uw}
\end{equation}

A comparison between the results obtained employing CFD and those estimated by the empirical expressions (\ref{axial_drag}), (\ref{lift_z})-(\ref{lift_n}) is provided in Table \ref{sim_coeff_results}. The axial drag coefficient of the hull, $X_{{u|u|}_{\rm hull}}$, as obtained by CFD, is lower by only 5\%, compared with the value obtained from experimental data of ellipsoidal bodies, suggesting that the empirical estimation provides a reasonable result. When examining the axial drag of the whole vehicle, however, the coefficient $X_{u|u|}$ obtained by CFD is higher by 10\% than the value obtained by employing strip theory and superposition of basic shapes, implying that the interactions between the components as assumed by strip theory are not entire negligible.

The CFD computation of the cross-flow drag coefficients $Y_{v|v|}$ and $Z_{w|w|}$  is lower by up to 30\% from the empirical estimation. Similar results were achieved for the coefficients of the bare hull alone, $Y_{{v|v|}_{\rm hull}}$ and $Z_{{w|w|}_{\rm hull}}$, indicating that the main deviation between the CFD and the empirical estimation is attributable to inaccurate estimation of the bare hull coefficients.

The moment coefficients $M_{w|w|}$, $N_{v|v|}$, $M_{q|q|}$ are lower by up to 80\% than the coefficients estimated by empirical methods, suggesting that the typically employed empirical method for the estimation of the cross-flow drag employing strip theory may not be sufficiently accurate in this case, possibly because of the three-dimensional effect neglected by strip theory. The CFD obtained values for the hull lift coefficients  $Y_{uv}$ and $Z_{uw}$ are lower by only 10\% from the empirically based estimated values, providing a relatively good correlation between the empirically obtained and CFD values.

As the CFD computation was performed on the actual shape of the vehicle’s hull at its operational speeds, the CFD obtained values are considered a better approximation and were initially employed in the dynamic model.  

\begin{table}[t]\centering
	\begin{tabular}{ |c|c|c| }
		\hline	
		Coefficient              &   Analytic   &  CFD         \\  \hline
	    $X_{{u|u|}_{\rm hull}}$  &   -5.309     &  -5.008      \\  \hline
	    $X_{u|u|}$               &   -7.61      &  -8.375      \\  \hline
	    $M_{u|u|}$               &    0.144     &   0.9        \\  \hline	  
	    $Y_{{v|v|}_{\rm hull}}$  &   -190.265   &  -120.834    \\  \hline
	    $Y_{v|v|}$               &   -214.398   &  -146.95     \\  \hline
	    $K_{v|v|}$               &   -3.39      &  -6.5        \\  \hline
	    $N_{{v|v|}_{\rm hull}}$  &   -42.515    &  -16.851     \\  \hline
	    $N_{v|v|}$               &   -26.17     &  -12.675     \\  \hline	    
	    $N_{r|r|}$               &   -180.381   &  -26.377     \\  \hline
	    $K_{r|r|}$               &    0.753     &  0.558       \\  \hline	    
	    $K_{p|p|}$               &   -0.192     &  -0.287      \\  \hline	    
	    $M_{q|q|}$               &   -180.68    &  -34.551     \\  \hline
	    $Z_{{w|w|}_{\rm hull}}$  &   -190.265   &  -120.834    \\  \hline
	    $Z_{w|w|}$               &   -214.398   &  -174.525    \\  \hline	    
	    $M_{{w|w|}_{\rm hull}}$  &    42.515    &   16.851     \\  \hline
	    $M_{w|w|}$               &    26.63     &   11.4       \\  \hline	    
	    $Y_{uv}$                 &   -39.71     &  -35.428     \\  \hline
	    $Z_{uw}$                 &   -39.71     &  -35.428     \\  \hline
	    $M_{uw}$                 &   -8.498     &  -3.37       \\  \hline
	    $N_{uv}$                 &    8.498     &   3.37       \\  \hline
\end{tabular}
	\caption{Comparison Between Estimated and CDF Computed Coefficients}
	\label{sim_coeff_results}
\end{table}

\subsection{Propulsion System Modeling}

The propulsion forces and moments vector $\Vec{\tau_c}$ were estimated by modeling the thrusters’ propellers, the interaction between the thrusters and the hull, and the thrusters’ configuration matrix, which defines the contribution of each thruster to the propulsion forces and moments vector.
The axial (thrust) force $T$ and torque $Q$ created by a thruster’s propeller are defined by~\cite{carlton2018marine}: 
\begin{equation}\label{thrust}
T = \rho D^4 k_t n |n|
\end{equation}
\begin{equation}\label{torque}
Q = \rho D^5 k_q n |n|
\end{equation}
where $D$ is the propeller diameter, $n$ is the rotational speed, and $k_t$ and $k_q$ are the thrust and torque coefficients, respectively, for a specific propeller configuration. The thrust and torque coefficients are estimated using the experimental data provided in~\cite{carlton2018marine} and~\cite{van1957recent}.

\subsection{Interactions Between the Horizontal Thrusters and the Hull}

The horizontal starboard and port side thrusters, denoted respectively by $T_{\rm th1}$ and $T_{\rm th2}$, are externally mounted on the hull and, therefore, interact with the wake field. Consequently, the speed of advance of the propeller relative to the water $u_p$ is less than the speed of the hull. The ratio between the speed of the hull and the speed of advance of the propeller is given by the wake fraction coefficient $w_T$:

\begin{equation}\label{wake_factor}
w_T = \frac{u-u_p}{u}
\end{equation}

In addition, the suction due to the operation of the propeller reduces the local pressure behind the hull, thus increasing the vehicle’s axial drag. Denoting the total thrust produced by the horizontal thrusters by $T_{\rm th1,2}$, this effect is modeled by the thrust deduction factor $t$:

\begin{equation}\label{thrust_reduction}
t =\frac{ T_{\rm th1,2} - R}{T_{\rm th1,2}}
\end{equation}

where $R$ is the total resistance of the bare hull. The wake fraction coefficient and the thrust deduction factor are incorporated into the dynamic model and estimated following empirical data for a hull with a twin-screw propeller configuration as provided in~\cite{burcher1995concepts}.

\subsection{Interactions Between the Lateral and Vertical Thrusters and the Hull}

\begin{figure}[t]
  \includegraphics[width=3.5in]{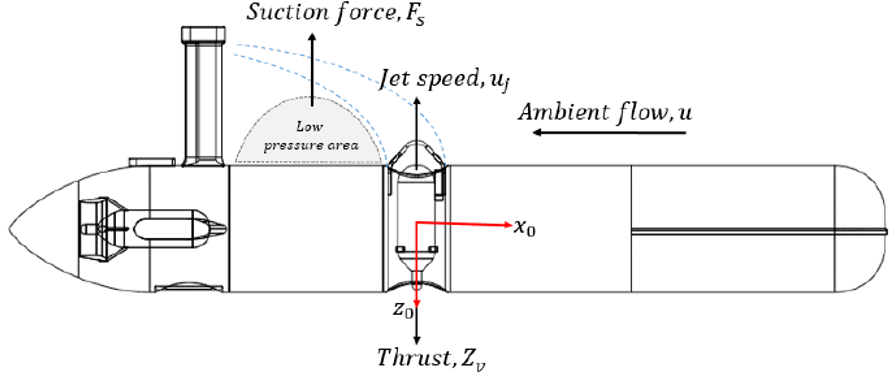}
  \caption{Interaction between the jet of the vertical tunnel thruster and the ambient flow.}\label{jet_deflection}
\end{figure}

The vertical and lateral nose and tail tunnel thrusters, denoted, respectively, by $th_3, th_4, th_5$, are considered jet producing devices. When the vehicle moves with a surge speed, the flow around the hull deflects the jets, creating a low-pressure area ~\cite{palmer2009analysis} as illustrated in Fig.~\ref{jet_deflection}. As a result, due to the offset between the center of the low pressure area and the thruster axis, the net thrust is reduced and a moment is created. Assuming the frictional effects due to the interactions between the jet and the tunnel are negligible and considering the effects of the jet deflection, the thrust created by the tunnel thrusters is expressed by~\cite{palmer2009analysis}:

\begin{equation}\label{thrust345}
T_{th_{3,4,5}} = \rho D^4 k_t e^{-C(\frac{u}{u_j})^2} n |n|
\end{equation}

where $C$ is a thrust deduction factor and $u_j$ is the jet speed, given by:

\begin{equation}\label{jet_speed}
u_j = \sqrt{\frac{T_{0}}{\rho A_{\rm th}}}
\end{equation}

where $T_{0}$ is the open water thrust as given by (\ref{thrust}) and $A_{th}$ is the cross-section area of the tunnel.

The jet developed by the tunnel thrusters interacts with the flow along the hull, creating an additional drag, which may be modeled \cite{tanakitkorn2017depth} by:

\begin{equation}\label{jet_drag}
X_{u|u|}^{\rm th} = - \frac{1}{2} \rho \nabla ^{\frac{2}{3}} C_d^{\rm th}
\end{equation}
where $\nabla$ is the volumetric displacement of the vehicle and $C_d^{\rm th}$ is the volumetric drag coefficient, estimated by employing experimental results of a vehicle with similar configuration ~\cite{tanakitkorn2017depth}.

\subsection{Thruster Configuration Matrix}
The contribution of the thrust created by each thruster to the force and moment vector, $\Vec{\tau_c}$ is defined by:
\begin{equation}\label{tcm}
\Vec{\tau_c} = B \Vec{f}_{\rm th}
\end{equation}
where $\Vec{f}_{\rm th} = [T_{\rm th_1}, T_{\rm th_2}, T_{\rm th_3}, T_{\rm th_4},T_{\rm th_5}]^T \in \Re^n $ is a column vector, representing the thrust created by each thruster and $B \in \Re^{6x5}$ is the thruster configuration matrix, given by: 
\begin{equation}\label{tcm_matrix}
B = 
\begin{pmatrix}
1       &  1         & 0   & 0         & 0        \\ 
0       &  0         & 0   & 1         & 1        \\
0       &  0         & 1   & 0         & 0        \\ 
0       &  0         & 0   & 0         & 0        \\ 
0       &  0         & 0   & 0         & 0        \\
y_{\rm th}  &  -y_{\rm th}   & 0   & x_{\rm th4}   & x_{\rm th5}  \\ 
\end{pmatrix} 	
\end{equation}
where the distances $y_{\rm th}$, $x_{\rm th4}$, $x_{\rm th5}$ are defined in Fig.~\ref{sparus_dimensions}.

\subsection{Numerical Implementation of the Dynamic Model}

The differential equations of motion, presented in~(\ref{general_form}), are highly nonlinear and coupled. To compute the acceleration, velocity and orientation of the vehicle, equation~(\ref{general_form}) is rearranged and solved numerically:
  
\begin{equation}\label{numeric_form}
\dot{\Vec{\nu}}_{(t)}  = M^{-1} \big[\Vec{\tau_c}_{(t)} - C(\Vec{\nu}_{(t)})\Vec{\nu}_{(t)} - D(\Vec{\nu}_{(t)} )\Vec{\nu}_{(t)}  - \Vec{g}(\Vec{\eta}_{(t)} ) \big]
\end{equation}

Equation~(\ref{numeric_form}) was implemented in the \emph{Matlab-Simulink \footnote{\url{https://www.mathworks.com/products/simulink.html}}} environment, solving $\dot{\Vec{\nu}}, \Vec{\nu}$ for the input $\tau_c$, where $\Vec{\nu}$ is solved by numeric integration of $\dot{\Vec{\nu}}$ and the vehicle's trajectory and orientation, referenced to the Earth-fixed frame, $\Vec{\eta}$ is obtained by numeric integration of (\ref{transformation}).

\section{Model Calibration and Validation in Real Sea Experiments}\label{Model_validation}

The ALICE AUV underwent sea trials off the coast of Sdot Yam, Achziv and in Haifa Bay. The vehicle’s motion during the experiments was measured by the on-board sensors, where the velocities  $[u, v, w]$ were measured by DVL, the angular rates $[p,q,r]$ by IMU and the angles  $[\psi,\theta,\phi]$ were computed by the vehicle’s angle estimator algorithm. The depth $Z$ was measured by a pressure sensor. For calibration and validation of the previously developed dynamic model, a hybrid approach (vehicle - dynamic model) was employed. The real-time dependent vehicle thrusters’ RPM commands as recorded in the dedicated sea trials were fed as input to the dynamic model and the simulated vehicle behavior as obtained from the dynamic model was compared with the real one as recorded during the sea trials.

Two recorded sea trial datasets were used for the two-step calibration–validation process. The first set was employed in the first step for hydrodynamic parameter calibration of the dynamic model. For the second step, the second dataset was employed with the goal of validating the previously calibrated model. The first step (calibration) was performed in a decoupled fashion (vertical and horizontal planes) by comparing the computed dynamics with the vehicle motions as recorded in dedicated sea experiments for the following maneuvers:

\begin{enumerate}
    \item Forward motion at various surge speeds to determine $X_{\dot{u}}$ and $X_{u|u|}$.
    \item Vertical ascent/descent at a constant heave speed to determine $Z_{\dot{w}}$ and $Z_{w|w|}$.
    \item Horizontal turns, performed by the horizontal thrusters, to determine $N_{\dot{r}}$ and $N_{r|r|}$.
\end{enumerate}

To differentiate the thruster-produced forces from the hydrodynamic forces in the calibration process, the thrust coefficients (and thus the thruster-produced forces) were initially calibrated in a bollard pull test while the hydrodynamic coefficients were iteratively adjusted to fit the vehicle’s measured response in each sea experiment. In particular, the added mass terms were calibrated according to recorded accelerations where the coefficients related to hydrodynamic damping were calibrated according to recorded velocities. A list of the coefficients calibrated by employing the experimentally obtained datasets is presented in Table~\ref{coeff_exp}.

The vehicle dynamics in surge was calibrated employing recorded data of the vehicle moving along straight transects at surge speeds between 0.3 to 1.5 m/s. The dynamic model presents sensitivity to small deviations between the horizontal thrusters’ rotational speeds leading to a yaw motion, which is barely observed in the experimental data. This difference between the model and the experiments may be a result of some existing differences between the thrust produced by the starboard side thruster and the port side thruster or by some external disturbances that are not modeled. To reduce this sensitivity in the dynamic model, the axial added mass coefficient $X_{\dot{u}}$ was calibrated by a factor of 10. Nevertheless, to better understand the reasons for the difference between the model and the experimental results, additional experiments are required. 

\begin{figure*}[t]
 \centering
  \includegraphics[width=1.0\linewidth]{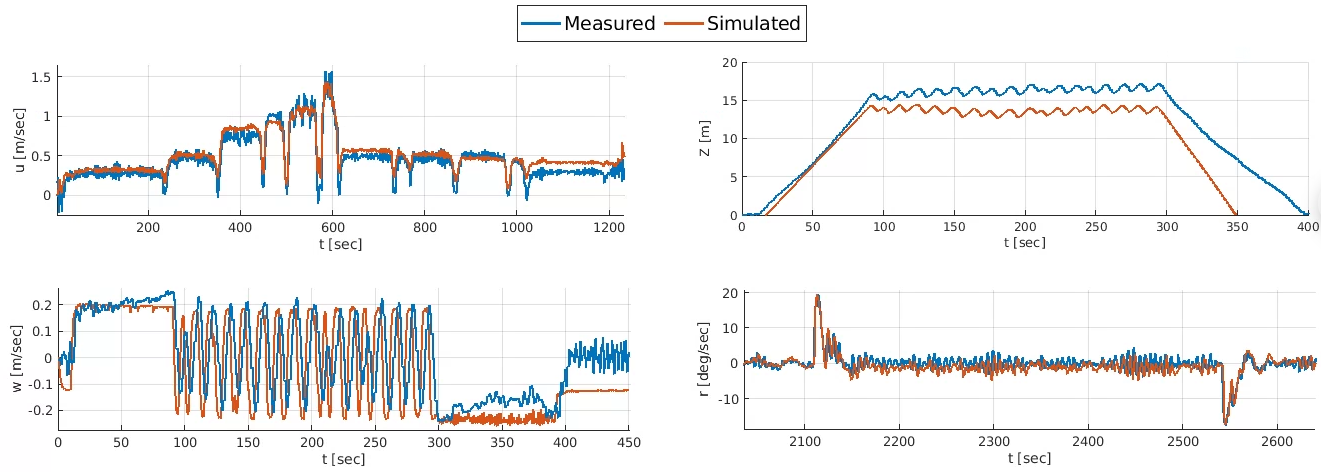}
  \caption{Model calibration results according to experimental data for surge (top left), heave (bottom left), depth (top right) and yaw rate (bottom right).}
  \label{calib_plots}
\end{figure*}

Moreover, the dynamic model surge velocities presented high values when compared with the experimental values. As a result, the axial hydrodynamic damping $X_{u|u|}$  was calibrated to match the measured velocities by a factor of 1.8. The error in the initial estimation of the axial drag coefficient may be attributed to various appendages and cavities on the hull, such as the acoustic modem, camera window, strobes and the gaps between the hull’s sections, which are not included in the CFD model. These inconsistencies, albeit relatively small in dimension, may have an extremely significant effect on the drag. 
In addition, some existing sea currents may have caused an additional error as the measured velocities were DVL based, thus relative to the seabed, while the model solves the vehicle’s speed relative to the water. The comparison between the simulated and the measured velocity in surge is presented in the top left plot of Fig.~\ref{calib_plots}. It can be noted that a reasonable fit is achieved; however, at time intervals t = 360-440 s, t = 620-730 s and t = 1030-1200 s, the dynamic model based computed velocities deviated from the measured velocities. The deviation between the measured and the simulation results are observed mainly when the vehicle flies along transects pointed in a specific direction. This may indicate that a significant current was acting in that specific direction, causing the observed discrepancy in the computed vs. measured velocity values. For further investigation, DVL data in water tracking mode may be employed. Such data, however, are noisy and require additional processing and filtering. Alternatively, a dedicated pool experiment may be conducted.

The calibration of the vehicle’s dynamics in heave was performed by employing a recorded dataset for a vertical dive to 15 meters, maintaining constant depth for 200 seconds followed by vertical ascent to the surface. Here, the coefficient $Z_{\dot{w}}$ was calibrated by a factor of 0.4 to fit the dynamic behavior in the vertical plane, the cross-flow drag coefficient, $Z_{w|w|}$ was calibrated by a factor of 1.86 to match the measured heave velocity and the moment coefficient $M_{w|w|}$ was calibrated by a factor of 0.23 to match the measured pitch angles. Similar to the axial drag case, it may be expected that the actual hydrodynamic drag of the hull is higher than the computed value due to appendages not considered in the CFD model. The comparison between the simulated and the measured heave speeds and the comparison between the measured and simulated depths presented, respectively, in the bottom left and top right plots of Fig.~\ref{calib_plots}, show excellent correlation between the simulated and measured velocities and depths during the dive (t = 0-75 s) and the depth keeping mode  period (t = 75-290 s). The computed heave velocity during the ascent to the surface, however, was slightly higher than the measured velocity, implying that the hull dynamics is not symmetrical for both heave directions. Additional heave maneuver experiments are required to determine the exact hull dynamics in each direction of motion.

The hydrodynamic coefficients associated with the dynamics in the horizontal plane were calibrated based on measured yaw rates and sway velocities during horizontal turns. To simplify the calibration process, a dataset of horizontal turns controlled only by the horizontal thrusters was employed. The added mass coefficient $N_{\dot{r}}$  was calibrated by a factor of 0.5 and $N_{\dot{v}}$ was calibrated by a factor of 0.3 to match the angular accelerations while the hydrodynamic damping moment coefficient $N_{r|r|}$  was calibrated by a factor of 0.3 to match the measured angular rates during the horizontal turns. The comparison between the measured values and the simulation results, presented in the bottom plot of Fig.~\ref{calib_plots}, shows excellent correlation between the model and the measured data and suggests successful calibration of the coefficients associated with the vehicle dynamics in the horizontal plane.

It should be noted that also in \cite{cardenas2019estimation} and \cite{de2020self} differences of the same order of magnitude were observed between values of coefficients computed (empirically or by CFD) and those obtained from analysis of real sea experiments. This underscores the need for pool tests or real sea calibration of dynamic models prior to their employment in scenario simulation or in development of propulsion control systems.

\begin{figure*}
  \centering
  \includegraphics[width=1.0\linewidth]{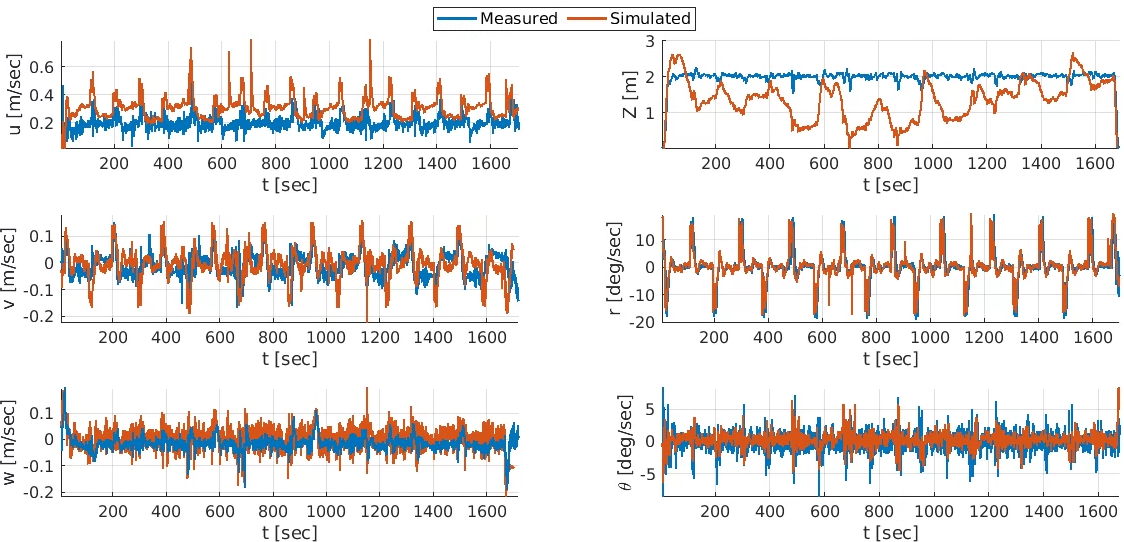}
  \caption{comparison between the vehicle's dynamic response, as computed by the calibrated model to the actual response, as measured at sea for surge (top left), sway (middle left), heave (bottom left), depth (top right), yaw rate (middle right) and pitch (bottom right).}
  \label{final_plots}
\end{figure*}

For the second step in the calibration–validation process, the previously calibrated dynamic model was fed with a second set of experimental data recorded during a navigation between way-points in a lawn mower pattern at a surge velocity of 0.2 m/s and at a depth of 2 m. The comparison between the simulation results and the measured surge velocity, presented in the top left plot of Fig.~\ref{final_plots}, shows good correlation between the measured and computed surge velocities, where a velocity of 0.2 m/s was computed when the vehicle flies along transects pointed South (e.g. at t= 120-200 s, t=300-380 s), however, when the vehicle flies along transects pointed North, the simulation solved a surge velocity of 0.3 m/s. This may indicate that a significant current was acting in that specific direction. A surge velocity of 0.3–0.5 m/s was observed and computed during the horizontal turns, when propelled by the horizontal thrusters. This significant surge velocity created by the horizontal thrusters (in additional to the angular yaw velocity) when maintaining identical rotational velocities but in the opposite direction was probably caused by difference in the thrust produced by the thrusters and thus their efficiency in forward versus astern directions.

The comparison between the simulation and measured sway velocities, presented in the middle left plot of Fig.~\ref{final_plots}, presents good correlation, in particular during the horizontal turns at t = 200 s and t = 300 s, where a sway velocity of 0.15 m/s was computed and measured. Nevertheless, a small deviation of about 0.05 m/s was observed during the transects, suggesting that the vehicle was subjected to a cross current resulting in a side-slip angle.

The comparison between the simulation and measured heave velocities, presented in the bottom left plot of Fig.~\ref{final_plots}, shows excellent correlation, in particular during the dive at t = 10 s, where a velocity of 0.18 m/s was computed and measured and during the depth keeping period, at t = 25-1700 s where a velocity around zero was computed and measured. The velocity computed during the ascent to the surface, however, was 13\% higher than the measured velocity. Additional validation of the vehicle’s dynamics in heave is shown in the top right plot of Fig.~\ref{final_plots}. The simulated depth, computed by integration of the heave velocity over the simulation time, did not exceed 1.5 m for an extended time period (over 1680 seconds). In addition, good correlation is observed in the comparison between the simulation and measured pitch angles, shown in the bottom right plot of Fig.~\ref{final_plots}, confirming that the simulation is capable of accurately predicting the vehicle’s dynamics in the vertical plane.
Considering the dynamics in the horizontal plane, the comparison between the simulation and measured yaw rates, shown in the middle right plot of Fig.~\ref{final_plots}, shows good correlation, in particular during the horizontal turns at time t = 120 s and t = 200 s. An identical yaw rate of 17 $^\circ$/s was measured and computed.
Following the successful calibration and validation process, the developed model was employed in the fine-tuning process of the vehicle’s control system and of the auto pilot system.

\begin{table}\centering
	\begin{tabular}{ |c|c|c|c| }
		\hline
		Coefficient   & Model   &   Experiment   &  Calibration factor \\  \hline
	    $X_{\dot{u}}$ &  -2.806    &   -28.06    &   10    \\  \hline
	    $Y_{\dot{v}}$ & -78.459    &   -23.53    &   0.3   \\  \hline
	    $Y_{\dot{r}}$ & -8.529     &   -2.559    &   0.3   \\  \hline
	    $Z_{\dot{w}}$ & -69.536    &   -27.812   &   0.4   \\  \hline
	    $N_{\dot{v}}$ & -8.529     &   -2.558    &   0.3   \\  \hline
	    $M_{\dot{w}}$ & -11.253    &   -5.626    &   0.5   \\  \hline
	    $M_{\dot{q}}$ & -20.963    &   -10.481   &   0.5   \\  \hline
	    $N_{\dot{r}}$ & -22.537    &   -11.268   &   0.5   \\  \hline
	    $X_{u|u|}$    & -8.375     &   -15.23   &   1.8   \\  \hline
	    $Y_{v|v|}$    & -146.95    &   -321.597  &   2.2   \\  \hline
	    $Z_{w|w|}$    & -174.52    &   -326.169  &   1.86  \\  \hline
	    $M_{w|w|}$    & 11.4       &    0.096    &   0.008  \\  \hline
	    $N_{v|v|}$    & -12.67     &   -1.954   &   0.15   \\  \hline
	    $N_{r|r|}$    & -26.377    &   -54.114   &   2.05  \\  \hline
\end{tabular}
	\caption{Comparison Between Estimated and Calibrated Coefficients.}
	\label{coeff_exp}
\end{table}

\section{Thruster Allocation Algorithm}\label{thruster_alocation}

For the ALICE AUV, the yaw motion is an over-actuated DOF since it can be controlled by both the horizontal and lateral thrusters. To compute the optimal thrust allocation, the redistributed pseudo-inverse (RPI) method~\cite{johansen2013control},~\cite{khan2018robust} was implemented. The method takes into consideration the thrusters’ availability in terms of thrust saturation, thus providing improved yaw control and redundancy in case of thruster failure. Employing the RPI method, the control vector $\Vec{f_{\rm th}}$ was iteratively computed by solving~(\ref{rpi_method}):

\begin{equation}\label{rpi_method}
\Vec{f_{th}} = -c + W^{-1} B^T \big[ B W^{-1} B^T + \epsilon I_3 \big]^{-1} \big[ \tau_c + B_0 c \big]
\end{equation}

where $B_0$ is the unconstrained allocation matrix (\ref{tcm_matrix}), $B$ is the modified allocation matrix, $W$ is weighting matrix, $c$ is the thruster saturation vector, $I_3$ is an identity matrix and $\epsilon$ is a small regularization parameter, defined to avoid singularity in the pseudo-inverse expression when $B$ does not possess full rank.
The elements of $B$ are initialized with $B_0$ and updated according to the thrusters’ availability while the vector $c$ is initialized with zeros and updated according to the thrusters’ saturation. The elements of the weighting matrix $W$ may be defined and updated in real time according to the vehicle and thruster status. For ALICE, at high surge speeds where the efficiency of the lateral thrusters is low, the horizontal thrusters are preferred over the lateral thrusters, while at slow speeds and, in particular, during imaging surveys, the torque produced by the horizontal thrusters may cause an undesired roll motion and, therefore, in this situation, the lateral thrusters are preferred. This logic is implemented in the weighting matrix.

\section{Improved Path-Following Controller}\label{path_controller}

\begin{figure}[t]
  \includegraphics[width=3.5in]{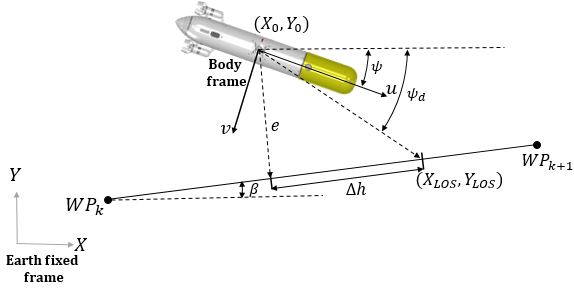}
  \caption{LOS controller geometry}\label{los}
\end{figure}

The under-actuated original configuration of the SPARUS II vehicle employed the popular line-of-sight (LOS) controller~\cite{palomeras2018auv} to follow a straight line, created by way-points where the following geometric computation is performed to find the required heading toward an interception with a moving point on the track:

\begin{equation}\label{los_heading}
\psi_{\rm LOS} = \tan^{-1}\bigg(\frac{Y_{\rm LOS} - Y_o}{ X_{\rm LOS} - X_o}\bigg)
\end{equation}

where $(X_o, Y_o)$ are the coordinates of the vehicle and $(X_{\rm LOS}, Y_{\rm LOS})$ are the coordinates of the moving point of interception determined by the look-ahead distance, $\Delta h$, as illustrated in Fig.~\ref{los}. Ideally, the cross-track error $e$ will converge to zero and the vehicle’s heading will converge to the direction of the path $\beta$. When lateral disturbances such as cross currents are present, however, lateral drift is avoided by steering the vehicle in a side-slip angle, which results in an undesired heading and may result in poor path following due to slow response by the indirect motion control. Consequently, this behavior impairs the main mission of the AUV – to perform high quality imaging surveys.
Considering the upgraded configuration of our ALICE AUV, the additional thrusters are able to create direct thrust to compensate for the drift, thus eliminating the necessity of crawling at a side-slip angle and removing the undesired roll due to torque created when differential thrust is applied.

The direct sway motion control provides the ability to effectively respond to lateral disturbances while maintaining the heading aligned with the direction of the path, thereby improving the accuracy of the path following and, consequently, the results of the visual survey. The sway motion control is incorporated into the LOS controller to nullify the lateral component of the cross-track error $e_{v}$, as defined by:

\begin{equation}\label{los_sway_e}
e_{v} = e \cdot \cos(\beta - \psi)
\end{equation}
and a PID controller is designed to control the sway motion as follows:   
\begin{equation}\label{los_sway_control}
v_{\rm req} = k_P \cdot e_{v} +  k_I \cdot \int e_{v} dt + k_D \cdot \dot{v} 
\end{equation}
where $k_p$, $k_i$ and $k_d$ are, respectively, the proportional, integral and derivative gains.

The sway motion is, however, considered inefficient due to the vehicle’s high lateral drag compared to the axial drag dominating the surge motion. Therefore, a switching strategy between the classic, heading control law and the proposed sway motion controller is implemented such that for a small cross-track error the vehicle follows the path laid out by the sway controller presented in (\ref{los_sway_control}) and maintains the path direction ($\psi_{\rm req} = \beta$) while for a large cross-track error, the vehicle will converge to the path dictated by the heading control law as presented in (\ref{los_heading}). 

\section{Real Sea Experiments of the Improved Path Following Controller}\label{imaging_experiments}

To evaluate the vehicle performance and examine the contribution of the new configuration in imaging missions, several surveys were conducted over Israel’s Mediterranean seabed in coastal areas. The modified path-following controller and thruster allocation algorithm were employed for controlling the vehicle while the visual imaging payload collected high resolution images at an altitude of 2 m above the seabed. The collected data was processed using the \emph{Agisoft Metashape\footnote{\url{https://www.agisoft.com/}}} commercial 3D photogrammetry software to produce large-scale mosaics and 3D reconstruction models.

\subsection{Mission Planning for High-Resolution Imaging}
\label{mapping_mission}

The imaging mission was planned following the guidelines described in~\cite{kwasnitschka2016deepsurveycam}. The imaging transects are defined by way-points in a lawnmower pattern consisting of equally spaced one-meter straight transects at an altitude of 2.0 m and a surge speed of 0.2 m/s. These settings provided an overlap of $60\%$ along the transects (in the propagation direction) and $45\%$ between two adjacent transects. The camera parameters were set according to the altitude, speed and visibility conditions. To avoid motion blur at velocities up to $0.5$ m/s, the exposure period was set to $3$ ms. The aperture was set to $F/4.0$ to optimize the amount of light entering the camera lens for the desired depth-of-field of $2$ m, while the cameras were focused to a distance of $2$ m in the water ($1.5$ m in the air). The camera frame rate $r$ was set according to~(\ref{fps_clc}): 

\begin{equation}\label{fps_clc}
    r =  \frac{u}{H \cdot (100\%-O)} \;\;
\end{equation}

where $u$ is the surge speed, $H$ is the altitude above the seabed and $O$ is the desired overlap, in percentages. 
 
The camera gain was set according to the visibility conditions in the surveyed site. To determine the gain, the vehicle performed a short dive prior to the survey mission and control images were collected. Once the vehicle surfaced, the gain was evaluated and corrected accordingly. 

\subsection{Evaluation of the Improved Path-Following Controller and Thruster Allocation}\label{path_experiments}

\begin{figure*}[t]
  \includegraphics[width=1.0\linewidth]{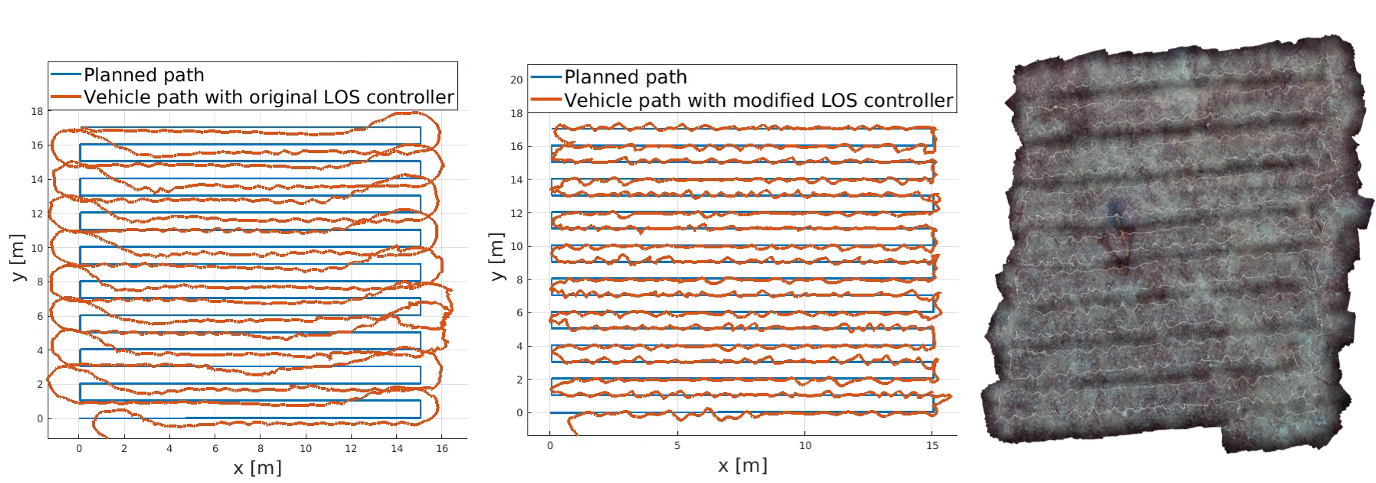}
  \caption{[Left]~vehicle's path, employing the original LOS controller along  a survey mission. [Center]~ vehicle's path, employing the modified LOS controller along  a survey mission. [Right]~Actual path calculated from camera positions on top of the generated mosaic.}
  \label{mission_compare}
\end{figure*}

\begin{figure}[t]
  \includegraphics[width=0.8\linewidth]{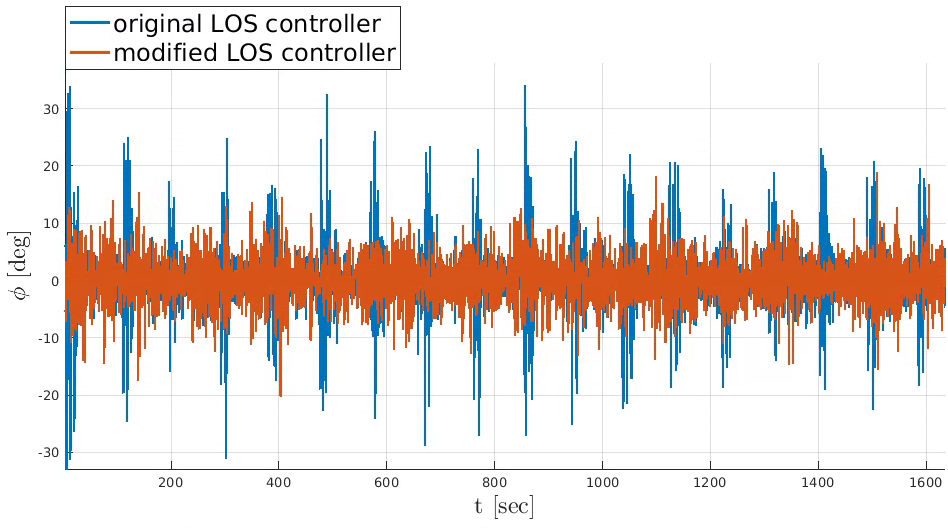}
  \caption{Comparison of the recorded roll angle, when employing the original and when employing the modified LOS controllers.}
  \label{mission_roll}
\end{figure}

\begin{figure*}
\begin{center}
\frame{\includegraphics[width=0.97\linewidth]{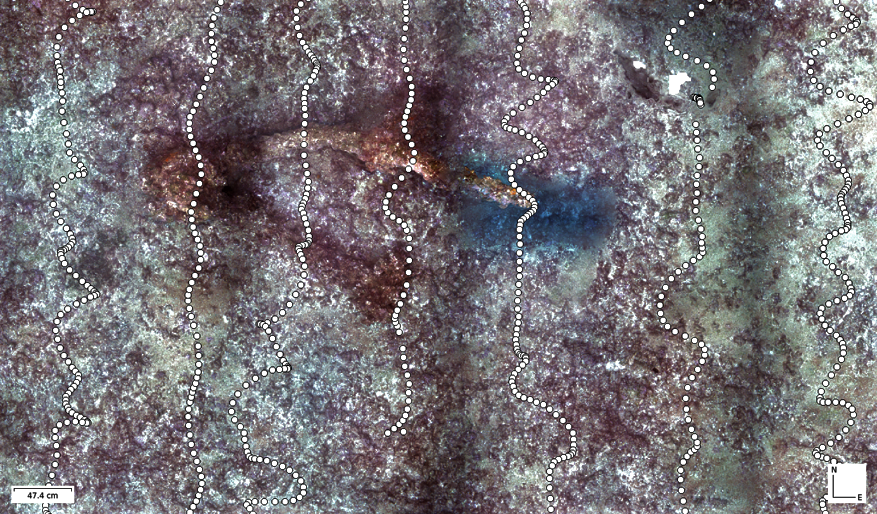}}
  \caption{Close up of the mosaic in Fig.~\ref{mission_compare}~[Right], acquired in Achziv, the north Israeli Mediterranean coast. It depicts a part of a shipwreck over a rocky reef. The AUV's path is depicted with white dots.}
    \label{ship_mosaic_track}
    \end{center}
\end{figure*}

\begin{figure}[t]
\begin{center}
 \frame{ \includegraphics[width=1.0\linewidth]{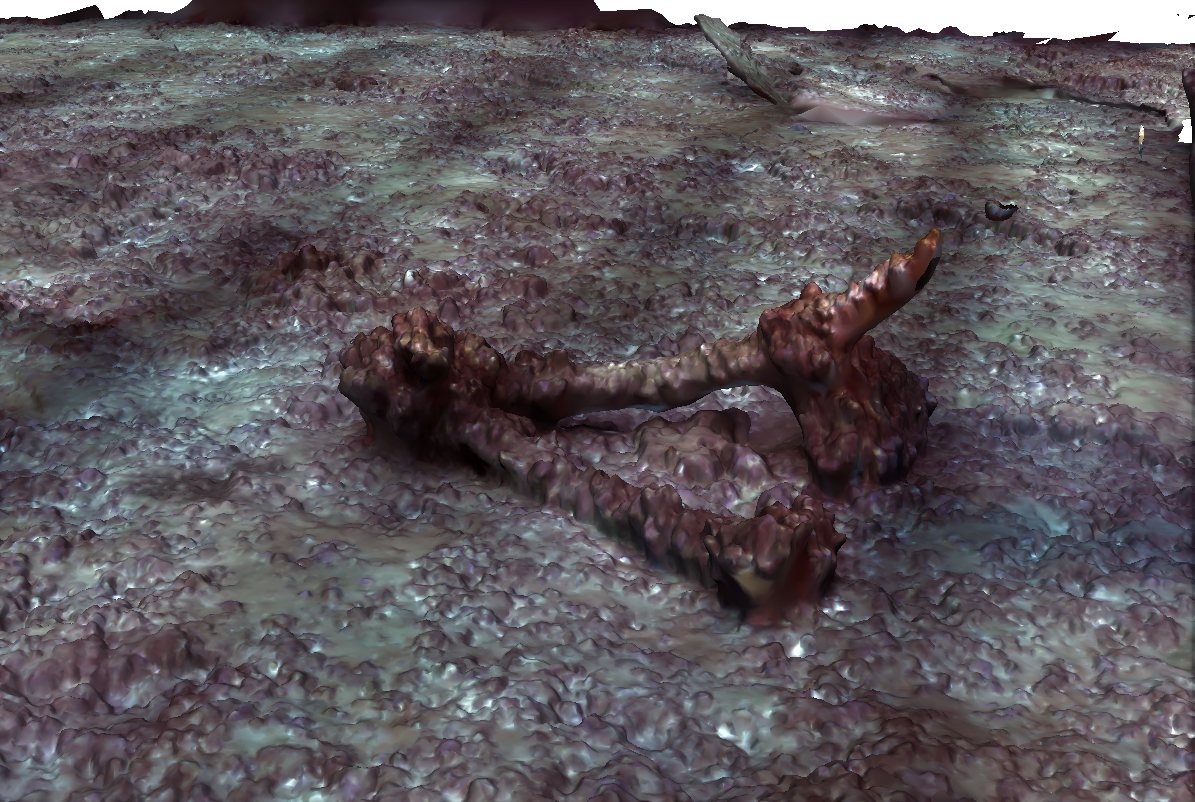}}
  \caption{Different viewpoint of the 3D reconstruction of the shipwreck shown in Fig.~\ref{ship_mosaic_track}.}
    \label{ship_mosaic_3d}
    \end{center}
\end{figure}

The performance of the new path-following controller was evaluated during a visual survey mission, performed twice, first with the upgraded controller and then again with the original controller. The survey site was selected to be a coastal area of $15 {\rm m} \times 17 {\rm m}$ around a part of a shipwreck. The vehicle was launched from a small support boat and initially performed the mission employing the upgraded controller. Following the completion of the first mission, the vehicle surfaced and the controller was switched to the original configuration, employing only the horizontal thrusters and the original LOS controller for path-following control. The vehicle path, employing the original configuration, is presented in Fig.~\ref{mission_compare}~[Left]. The results of the path following, performed with the original propulsion configuration, demonstrates the original vehicle’s inability to perform tight turns between the transects, resulting in slow convergence to the path and a root mean square cross-track error of $0.36$ m. The vehicle path, using the modified LOS controller, can be seen in Fig.~\ref{mission_compare}~[Center]. The results demonstrate a significant improvement in the path-following accuracy, where the path is precisely followed and a root mean square cross-track error of $0.12$ m is obtained, offering an improvement of $65\%$ in path-following accuracy.
In addition, a comparison between the vehicle’s roll angles as recorded during the two missions is presented in Fig.~\ref{mission_roll}. The results demonstrate the improvement in the roll stability due to the new thruster allocation, where the roll angle during the horizontal turns decreased from $20^o-30^\circ$ to less than $10^\circ$. The data collected during the mission performed with the modified controller was processed into a complete photomosaic and appears in Figs.~\ref{mission_compare}~[Right],~\ref{ship_mosaic_track} and~\ref{ship_mosaic_3d}, where the vehicle path, as computed by the Agisoft Metashape software, is marked by white dots in Fig~\ref{ship_mosaic_track}, providing an additional validation of the path-following accuracy.

\begin{figure*}
  \includegraphics[width=1.0\linewidth]{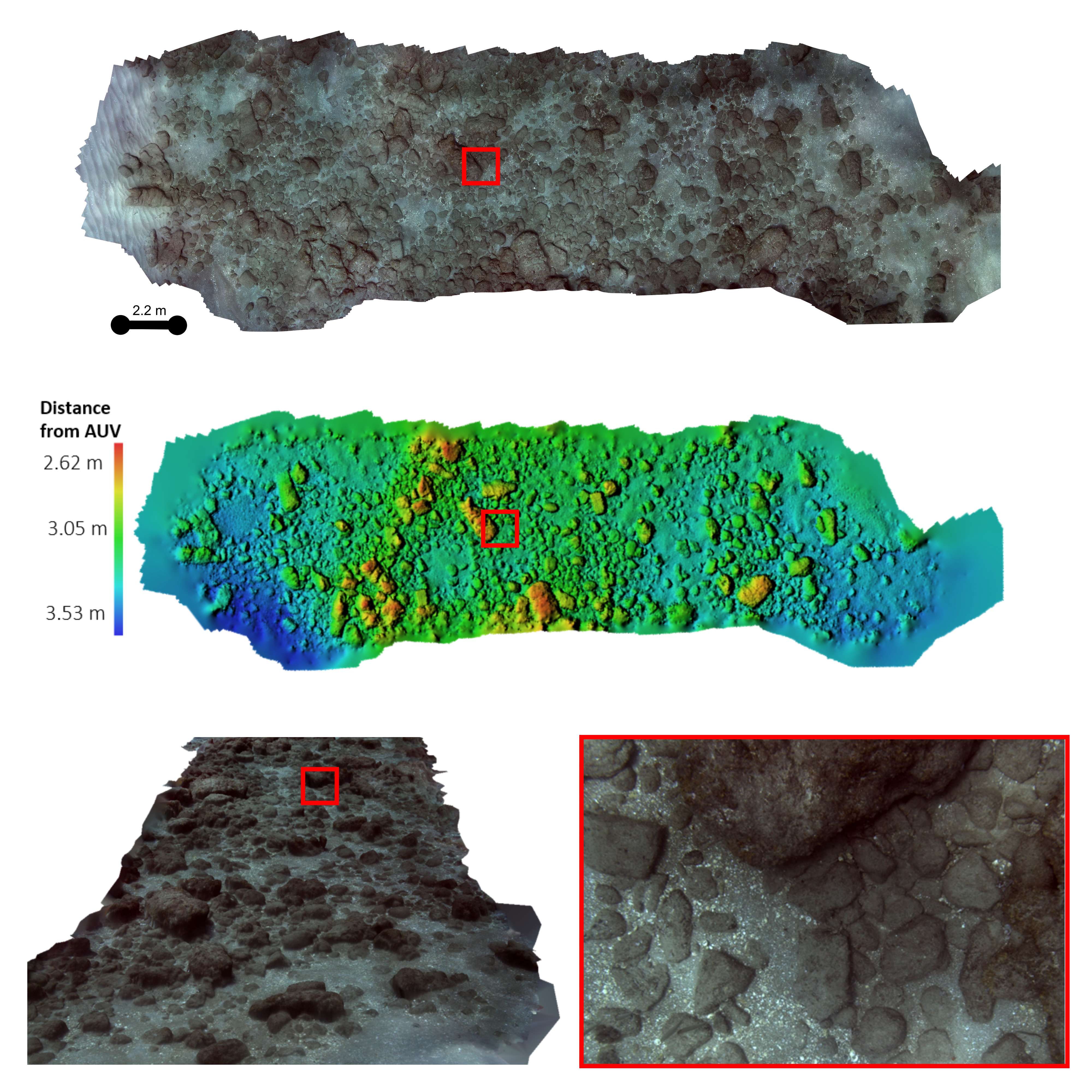}
  \caption{3D mosaic of a Roman Underwater Harbour, in Caesarea, Israel. [Top]~Mosaic of an area of $8{\rm m} \times 27{\rm m}$. [Middle]~3D reconstruction of the same area. [Bottom left]~Another viewpoint of the reconstructed 3D. [Bottom right]~Zoom in of an area marked with a red rectangle in the top mosaic. Full model can be viewed in \protect \url{https://sketchfab.com/3d-models/caesarea-roman-underwater-harbour-da2ca77fcf6141bb8e1261fb6f1d1187}.}
    \label{collage_med}
\end{figure*}

An additional survey mission was conducted as part of a collaboration with the marine division of the Israel Antiquities Authority with the goal of reconstructing the structure of the ancient harbor of Caesarea. The survey area was selected to be an area of $10 {\rm m}\times 30 {\rm m}$ at $8$m depth and included part of an ancient dock. The survey was performed at an altitude of $3$m to provide additional safety margins due to large rocks and unstructured seabed at the site. The results of the survey, presented in Fig.~\ref{collage_med}, demonstrate the current vehicle’s ability to collect high resolution images and provide the necessary data for precise 3D reconstruction.

\section{Conclusions} \label{Conclusions}

In this work, we reported on the modifications performed on a small sized AUV with given maneuverability so that it became an autonomous imaging platform for precise and high-resolution visual seabed mapping and inspection. Two lateral thrusters were added to enable sway and decoupled yaw control. An imaging system, consisting of stereo cameras and strobes, was developed and integrated into the vehicle. To study the vehicle maneuverability and develop the upgraded propulsion and control systems, a dynamic model was developed and validated with experimental data in a hybrid simulation process. An upgraded thruster allocation algorithm was implemented to optimize the vehicle’s dynamic stability in roll and enable decoupled motion control in yaw and sway. Also, an improved path-following algorithm was developed to increase the vehicle’s ability to follow the survey path accurately in real sea conditions. A comparison between the vehicle’s original configuration and the modified configuration demonstrated the improved performance. Finally, the ability of the modified AUV ALICE to perform imaging surveys was successfully established in a series of missions at sea, where imaging data was collected and then processed into 3D photomosaics.

\section{Acknowledgment} 

The authors would like to thank the IQUA robotics team for guidance and assistance with technical issues and integration and to Jacob Sharvit and Dror Palner from the Marine Archaeology Unit of the Israel Antiquities Authority. Special thanks to the support of NVIDIA Corporation through the donation of the Titan XP GPU used for 3D model generation. 

\ifCLASSOPTIONcaptionsoff
  \newpage
\fi




\bibliographystyle{IEEEtran}

\bibliography{IEEEabrv,Bibliography}

\begin{thebibliography}{10}
\providecommand{\url}[1]{#1}
\csname url@samestyle\endcsname
\providecommand{\newblock}{\relax}
\providecommand{\bibinfo}[2]{#2}
\providecommand{\BIBentrySTDinterwordspacing}{\spaceskip=0pt\relax}
\providecommand{\BIBentryALTinterwordstretchfactor}{4}
\providecommand{\BIBentryALTinterwordspacing}{\spaceskip=\fontdimen2\font plus
\BIBentryALTinterwordstretchfactor\fontdimen3\font minus
  \fontdimen4\font\relax}
\providecommand{\BIBforeignlanguage}[2]{{%
\expandafter\ifx\csname l@#1\endcsname\relax
\typeout{** WARNING: IEEEtran.bst: No hyphenation pattern has been}%
\typeout{** loaded for the language `#1'. Using the pattern for}%
\typeout{** the default language instead.}%
\else
\language=\csname l@#1\endcsname
\fi
#2}}
\providecommand{\BIBdecl}{\relax}
\BIBdecl
\renewcommand{\BIBentryALTinterwordstretchfactor}{4}

\bibitem{pizarro2017simple}
O.~Pizarro, A.~Friedman, M.~Bryson, S.~B. Williams, and J.~Madin, ``A simple,
  fast, and repeatable survey method for underwater visual 3\textsc{D} benthic
  mapping and monitoring,'' \emph{Ecol.Evol}, vol.~7, no.~6, pp. 1770--1782,
  2017.

\bibitem{tolimieri2008evaluating}
N.~Tolimieri, M.~E. Clarke, H.~Singh, and C.~Goldfinger, ``Evaluating the
  seabed \textsc{AUV} for monitoring groundfish in untrawlable habitat,''
  \emph{in Marine Habitat Mapping Technology for Alaska, J. R. Reynolds and H.
  G. Greene Eds., US: Alaska Sea Grant for North Pacific Research Board}, pp.
  129--141, 2008.

\bibitem{gracias2013mapping}
N.~Gracias \emph{et~al.}, ``Mapping the moon: Using a lightweight \textsc{AUV}
  to survey the site of the 17th century ship ‘la lune’,'' in
  \emph{\OCEANS}, (Bergen, Norway), 2013.

\bibitem{albiez2015flatfish}
J.~Albiez \emph{et~al.}, ``Flatfish-a compact subsea-resident inspection
  \textsc{AUV},'' in \emph{\OCEANS}, (Washington, DC, USA), 2015.

\bibitem{vidal2018online}
E.~Vidal, N.~Palomeras, and M.~Carreras, ``Online 3\textsc{D} underwater
  exploration and coverage,'' in \emph{IEEE/OES Autonomous Underwater Vehicle
  Workshop \textsc{(AUV)}}, 2018.

\bibitem{ludvigsen2007applications}
M.~Ludvigsen, B.~Sortland, G.~Johnsen, and H.~Singh, ``Applications of
  geo-referenced underwater photo mosaics in marine biology and archaeology,''
  \emph{Oceanogr}, vol.~20, no.~4, pp. 140--149, 2007.

\bibitem{jakuba2010high}
M.~V. Jakuba, O.~Pizarro, and S.~B. Williams, ``High resolution, consistent
  navigation and 3\textsc{D} optical reconstructions from \textsc{AUV}s using
  magnetic compasses and pressure-based depth sensors,'' in \emph{\OCEANS},
  (Sydney, Australia), 2010.

\bibitem{kim2009toward}
A.~Kim and R.~M. Eustice, ``Toward \textsc{AUV} survey design for optimal
  coverage and localization using the cramer rao lower bound,'' in \emph{IEEE
  OCEANS}, (Biloxi, MS, USA), 2009.

\bibitem{pizarro2003toward}
O.~Pizarro and H.~Singh, ``Toward large-area mosaicing for underwater
  scientific applications,'' \emph{IEEE J. Ocean. Eng}, vol.~28, no.~4, pp.
  651--672, 2003.

\bibitem{prados2012novel}
R.~Prados, R.~Garcia, N.~Gracias, J.~Escartin, and L.~Neumann, ``A novel
  blending technique for underwater gigamosaicing,'' \emph{IEEE J. Ocean. Eng},
  vol.~37, no.~4, pp. 626--644, 2012.

\bibitem{pyo2015development}
J.~Pyo, H.~Cho, H.~Joe, T.~Ura, and S.-C. Yu, ``Development of hovering type
  \textsc{AUV} “cyclops” and its performance evaluation using image
  mosaicing,'' \emph{Ocean Eng}, vol. 109, pp. 517--530, 2015.

\bibitem{bewley2015australian}
M.~Bewley \emph{et~al.}, ``Australian sea-floor survey data, with images and
  expert annotations,'' \emph{Scientific data}, vol.~2, no.~1, pp. 1--13, 2015.

\bibitem{houts2012aggressive}
S.~E. Houts, S.~M. Rock, and R.~McEwen, ``Aggressive terrain following for
  motion-constrained \textsc{AUVs},'' in \emph{IEEE/OES Autonomous Underwater
  Vehicles \textsc{AUV}}, (Southampton, UK), 2012.

\bibitem{helgason2012low}
B.~Helgason, L.~Leifsson, I.~Rikhardsson, H.~Thorgilsson, and S.~Koziel,
  ``{Low-speed modeling and simulation of torpedo-shaped AUVs.}'' in
  \emph{Proc. ICINCO(2)}, (Rome, Italy), 2012, pp. 333--338.

\bibitem{packard2010hull}
G.~E. Packard, R.~Stokey, R.~Christenson, F.~Jaffre, M.~Purcell, and
  R.~Littlefield, ``Hull inspection and confined area search capabilities of
  \textsc{REMUS} autonomous underwater vehicle,'' in \emph{\OCEANS}, (Seattle,
  WA, USA), 2010.

\bibitem{philips2013delphin2}
A.~Philips, L.~Steenson, E.~Rogers, S.~Turnock, C.~Harris, and M.~Furlong,
  ``\textsc{delphin2}: An over actuated autonomous underwater vehicle for
  manoeuvring research,'' \emph{Trans. Roy. Institution of Naval Architects
  Part A: Int. J. Maritime Eng.}, vol. 155, no.~A4, pp. 171--180, 2013.

\bibitem{wirtz2016iceshuttle}
M.~Wirtz and M.~Hildebrandt, ``\textsc{iceshuttle teredo}: an ice-penetrating
  robotic system to transport an exploration \textsc{AUV} into the ocean of
  jupiter’s moon europa,'' in \emph{Proc. 67th International Astronautical
  Congr (IAC)}, vol.~9, (Guadalajara, Mex.), 2016.

\bibitem{wang2019maneuverability}
X.~Wang and S.~Liang, ``Maneuverability analysis of a novel portable modular
  \textsc{AUV},'' \emph{Math. Problems in Eng.}, 2019.

\bibitem{allotta2016archaeology}
B.~Allotta, R.~Costanzi, A.~Ridolfi, M.~Reggiannini, M.~Tampucci, and
  D.~Scaradozzi, ``Archaeology oriented optical acquisitions through
  \textsc{MARTA AUV} during arrows european project demonstration,'' in
  \emph{\OCEANS}, (Monterey, CA, USA), 2016.

\bibitem{carreras2018sparus}
M.~Carreras, J.~D. Hern{\'a}ndez, E.~Vidal, N.~Palomeras, D.~Ribas, and
  P.~Ridao, ``\textsc{SPARUS II AUV} —a hovering vehicle for seabed
  inspection,'' \emph{IEEE Journal of Oceanic Engineering}, vol.~43, no.~2, pp.
  344--355, 2018.

\bibitem{quigley2009ros}
M.~Quigley, K.~Conley, B.~Gerkey, J.~Faust, T.~Foote, J.~Leibs, R.~Wheeler, and
  A.~Y. Ng, ``\textsc{ROS: AN OPEN-SOURCE ROBOT OPERATING SYSTEM},'' in
  \emph{Proc. ICRA workshop on open source software}, vol.~3, no. 3.2, (Kobe,
  Japan), 2009, p.~5.

\bibitem{xia2019improved}
Y.~Xia, K.~Xu, Y.~Li, G.~Xu, and X.~Xiang, ``Improved line-of-sight trajectory
  tracking control of under-actuated \textsc{AUV} subjects to ocean currents
  and input saturation,'' \emph{Ocean Engineering}, vol. 174, pp. 14--30, 2019.

\bibitem{treibitz2011flat}
T.~Treibitz, Y.~Schechner, C.~Kunz, and H.~Singh, ``Flat refractive geometry,''
  \emph{IEEE Trans. Pattern Analysis Mach. Intell.}, vol.~34, no.~1, pp.
  51--65, 2011.

\bibitem{fossen2011handbook}
T.~I. Fossen, \emph{Handbook of marine craft hydrodynamics and motion
  control}.\hskip 1em plus 0.5em minus 0.4em\relax Hoboken, NJ: Wiley, 2011.

\bibitem{hagist1965experimental}
R.~R. Miller and W.~M. Hagist, ``Experimental determination of the hydrodynamic
  mass of various bodies.'' Rhode Island Univ Kingston, Tech. Rep., 1965.

\bibitem{horner1965fluid}
S.~Horner, ``{Fluid dynamic drag, practical information on aerodynamic drag and
  hydrodynamic resistance},'' \emph{Midland Park, NJ: Hoerner Fluid Dyn.},
  1965.

\bibitem{ittc2002guidelines}
R.~P. ITTC, ``Guidelines: Testing and extrapolation methods:
  Resistance-uncertainty analysis, example for resistance test,'' \emph{ITTC
  Recommended Procedures and Guidelines, Procedure 7.5-02-02}, vol.~2, 2002.

\bibitem{uematsu1995effects}
Y.~Uematsu and M.~Yamada, ``Effects of aspect ratio and surface roughness on
  the time-averaged aerodynamic forces on cantilevered circular cylinders at
  high reynolds numbers,'' \emph{J. Wind Eng. Ind. Aerodynamics}, vol.~54, pp.
  301--312, 1995.

\bibitem{palmer2009analysis}
A.~R. Palmer, ``Analysis of the propulsion and manoeuvring characteristics of
  survey-style \textsc{AUVs} and the development of a multi-purpose
  \textsc{AUV},'' Ph.D. dissertation, Univ. Southampton, UK, 2009.

\bibitem{hoerner1975fluid}
S.~F. Hoerner and H.~V. Borst, ``Fluid-dynamic lift: practical information on
  aerodynamic and hydrodynamic lift,'' \emph{STIA}, vol.~76, p. 32167, 1975.

\bibitem{carlton2018marine}
J.~Carlton, \emph{Marine propellers and propulsion}.\hskip 1em plus 0.5em minus
  0.4em\relax Oxford, UK: Butterworth-Heinemann, 2018.

\bibitem{van1957recent}
J.~D. Van~Manen, ``Recent research on propellers in nozzles,'' \emph{Int.
  Shipbuilding Prog.}, vol.~4, no.~36, pp. 395--424, 1957.

\bibitem{burcher1995concepts}
R.~Burcher and L.~J. Rydill, \emph{Concepts in submarine design}.\hskip 1em
  plus 0.5em minus 0.4em\relax Cambridge, UK: Cambridge Univ. press, 1995,
  vol.~2.

\bibitem{tanakitkorn2017depth}
K.~Tanakitkorn, P.~A. Wilson, S.~R. Turnock, and A.~B. Phillips, ``Depth
  control for an over-actuated, hover-capable autonomous underwater vehicle
  with experimental verification,'' \emph{Mechatronics}, vol.~41, pp. 67--81,
  2017.

\bibitem{cardenas2019estimation}
P.~Cardenas and E.~A. de~Barros, ``Estimation of auv hydrodynamic coefficients
  using analytical and system identification approaches,'' \emph{IEEE J. Ocean.
  Eng.}, vol.~45, no.~4, pp. 1157--1176, 2019.

\bibitem{de2020self}
B.~J. de~Kruif and E.~Ypma, ``Self-propulsion parameter identification for
  control of marin’s \textsc{AUV},'' in \emph{Proc. IEEE/OES Autonomous
  Underwater Vehicles Symposium (AUV)(50043)}, 2020, pp. 1--6.

\bibitem{johansen2013control}
T.~A. Johansen and T.~I. Fossen, ``Control allocation—a survey,''
  \emph{Automatica}, vol.~49, no.~5, pp. 1087--1103, 2013.

\bibitem{khan2018robust}
H.~Z.~I. Khan, J.~Rajput, S.~Ahmed, M.~Sarmad, and M.~Sharjil, ``Robust control
  of overactuated autonomous underwater vehicle,'' in \emph{Proc. IEEE 15th
  International Bhurban Conf. Appl. Sci. Tech. (IBCAST)}, 2018, pp. 269--275.

\bibitem{palomeras2018auv}
N.~Palomeras, G.~Vallicrosa, A.~Mallios, J.~Bosch, E.~Vidal, N.~Hurtos,
  M.~Carreras, and P.~Ridao, ``\textsc{AUV} homing and docking for remote
  operations,'' \emph{Ocean Eng.}, vol. 154, pp. 106--120, 2018.

\bibitem{kwasnitschka2016deepsurveycam}
Kwasnitschka \emph{et~al.}, ``Deepsurveycam—a deep ocean optical mapping
  system,'' \emph{Sensors}, vol.~16, no.~2, p. 164, 2016.

\end{thebibliography}

\vfill


\end{document}